\documentclass[acmtog,nonacm]{acmart}

\DeclareMathOperator*{\argmin}{arg\,min}

\newcommand{\afterfigure}{\vspace{-1em}}

\newcommand{\mathcls}{[\mathsf{CLS}]}
\newcommand{\cls}{$\mathcls{}$\xspace}

\newcommand{\dinovit}{DINO-ViT\xspace}

\newcommand{\myparagraph}[1]{\vspace{0.15cm}\noindent{\bf #1}\hspace{0.05cm}}

\usepackage[leftmargin=2mm,rightmargin=0pt]{quoting}
\usepackage{adjustbox}
\usepackage{ifthen}
\usepackage{xspace}
\usepackage[normalem]{ulem}

\AtBeginDocument{%
  \providecommand\BibTeX{{%
    \normalfont B\kern-0.5em{\scshape i\kern-0.25em b}\kern-0.8em\TeX}}}

\setcopyright{acmcopyright}
\copyrightyear{2023}
\acmYear{2023}
\acmDOI{XXXXXXX.XXXXXXX}

\begin{document}

\title{Disentangling Structure and Appearance in ViT Feature Space}

\author{Narek Tumanyan}
\email{narek.tumanyan@weizmann.ac.il}
\orcid{0000-0003-2734-8480}

\author{Omer Bar-Tal}
\email{omer-bar.tal@weizmann.ac.il}
\orcid{0000-0003-1622-3674}

\author{Shir Amir}
\orcid{0009-0004-4453-6920}
\email{shiramiremail@gmail.com}

\author{Shai Bagon}
\orcid{0000-0002-6057-4263}
\email{shai.bagon@weizmann.ac.il}

\author{Tali Dekel}
\email{tali.dekel@weizmann.ac.il}
\orcid{0000-0003-3703-0783}
\affiliation{
  \institution{\newline Weizmann Institute of Science}
  \city{Rehovot}
  \country{Israel}
}

\begin{abstract}

We present a method for semantically transferring the visual appearance of one natural image to another. Specifically, our goal is to generate an image in which objects in a source structure image are ``painted'' with the visual appearance of their semantically related objects in a target appearance image. To integrate semantic information into our framework, our key idea is to leverage a pre-trained and fixed Vision Transformer (ViT) model. Specifically, we derive novel disentangled representations of structure and appearance extracted from deep ViT features. We then establish an objective function that splices the desired structure and appearance representations, interweaving them together in the space of ViT features. Based on our objective function, we propose two frameworks of semantic appearance transfer -- ``Splice”, which works by training a generator on a \emph{single and arbitrary} pair of structure-appearance images, and ``SpliceNet”, a \emph{feed-forward} real-time appearance transfer model trained on a \emph{dataset} of images from a \emph{specific domain}. Our frameworks do not involve adversarial training, nor do they require any additional input information such as semantic segmentation or correspondences. We demonstrate high-resolution results on a variety of in-the-wild image pairs, under significant variations in the number of objects, pose, and appearance. Code and supplementary material are available in our project page: \href{https://splice-vit.github.io}{splice-vit.github.io}.
\end{abstract}



\begin{teaserfigure}
\includegraphics[width=\textwidth]{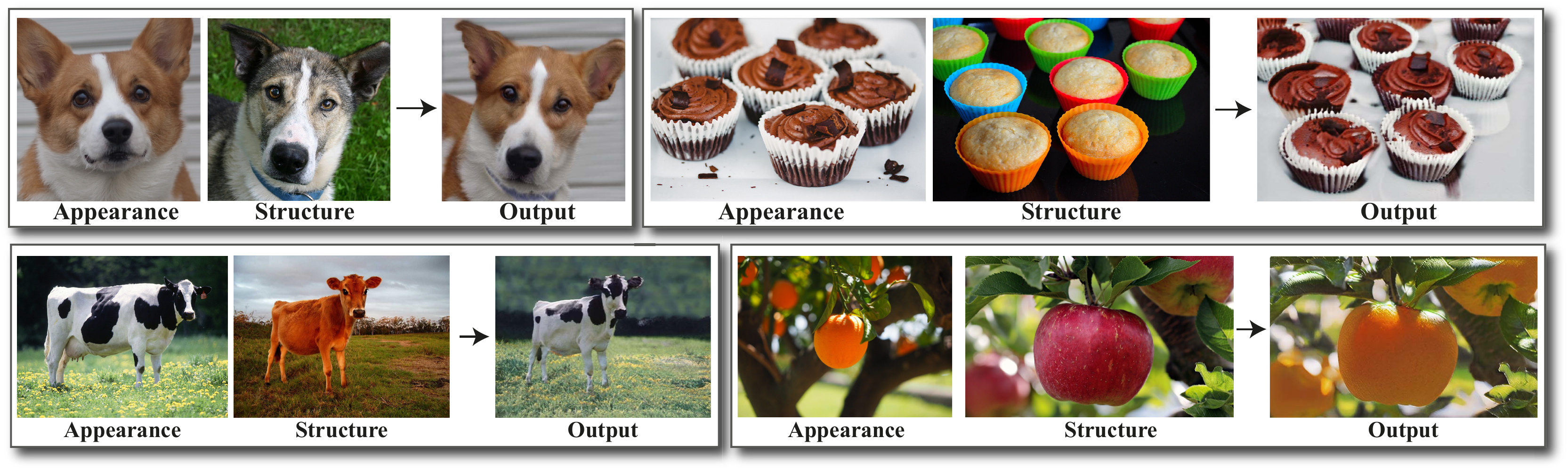}
\vspace*{-4mm}
\caption{
Given two input images--a source \emph{structure} image and a target \emph{appearance} image--our method generates a new image in which the structure of the source image is preserved, while the visual appearance of the target image is transferred in a \emph{semantically} aware manner. 
That is, objects in the structure image are ``painted'' with the visual appearance of semantically related objects in the appearance image. Our method leverages a self-supervised, pre-trained ViT model as an external semantic prior. We derive novel disentangled appearance and structure representations from our semantic prior, which allows us to train a generator without any additional information (e.g., segmentation/correspondences), and without adversarial training. Thus, our framework can work across a variety of objects and scenes, and can generate high quality results in high resolution (e.g., HD).
}
\vspace*{3mm}
\label{fig:teaser}
\end{teaserfigure}

\maketitle

\section{Introduction}
\begin{quoting}
\noindent\textit{``Rope splicing is the forming of a semi-permanent joint between two ropes by partly untwisting and then interweaving their strands.''~\cite{splicing}} 
\end{quoting}

\noindent What is required to transfer the visual appearance between two semantically related 
images? 
Consider for example the task of transferring the visual appearance of a spotted cow in a flower field to an image of a red cow in a grass field (Fig.~\ref{fig:teaser}). Conceptually, we have to associate regions in both images that are semantically related, and transfer the visual appearance between these matching regions. Additionally, the target appearance has to be transferred in a realistic manner, while preserving the structure of the source image~-- the red cow should be realistically ''painted'' with black and white spots, and the green grass should be covered with yellowish colors.  To achieve it under noticeable pose, appearance and shape differences between the two images, \emph{semantic} information is imperative.

Indeed, with the rise of Deep Learning and the ability to learn high-level visual representations from data, new vision tasks and methods under the umbrella of ``visual appearance transfer'' have emerged. For example, the image-to-image translation line of work aims at translating a source image from one domain to another target \emph{domain}. To achieve that, most methods use generative adversarial networks (GANs), given image collections from both domains. Our goal is different --  rather than generating \emph{some} image in a target domain, we generate an image that depicts the visual appearance of a \emph{particular} target image, while preserving the structure of the source image.

Given a pair of structure and appearance images, how can we source semantic information necessary for the task of semantic appearance transfer? We draw inspiration from Neural Style Transfer (NST) that represents content and an artistic style in the space of deep features encoded by a pre-trained classification CNN model (e.g., VGG). While NST methods have shown a remarkable ability to \emph{globally} transfer  artistic styles, their content/style representations are not suitable for \emph{region-based}, semantic appearance transfer across objects in two natural images~\cite{JingYFYYS20}.  Here, we propose novel deep representations of  appearance and structure that are extracted from DINO-ViT~-- a Vision Transformer model that has been pre-trained in a self-supervised manner~\cite{dino}. Representing structure and appearance in the space of ViT features allows us to inject powerful semantic information into our method and establish a novel objective function for semantic appearance transfer. Based on our objective function, we propose two frameworks of semantic appearance transfer: (i) a generator trained on a \emph{single and in-the-wild} input image pair, (ii) a \emph{feed-forward} generator trained on a dataset of \emph{domain-specific} images.

DINO-ViT has been shown to learn powerful and meaningful visual representation, demonstrating impressive results on several downstream tasks including image retrieval, object segmentation, and copy detection~\cite{dino, amir2021deep, melaskyriazi2022deep, LOST, wang2022tokencut}. However, the intermediate representations that it learns have not yet been fully explored.  We thus first strive to gain a better understanding of the information encoded in different ViT's features across layers. We do so by adopting  ``feature inversion'' visualization techniques previously used in the context of CNN features. Our study provides a couple of key observations: (i)~the global token (a.k.a \cls \ token) provides a powerful representation of visual appearance, which captures not only texture information but more global information such as object parts, and (ii) the original image can be reconstructed from these features, yet they provide powerful semantic information at high spatial granularity.

Equipped with the above observations,  we derive novel representations of structure and visual appearance extracted from deep ViT features -- untwisting them from the learned self-attention modules. Specifically, we represent visual appearance via the global  \cls \ token,  and represent structure via the self-similarity of keys, all extracted from the attention module of last layer. We then design a framework of training a generator on a \emph{single input pair} of structure/appearance images to produce an image that \emph{splices} the desired visual appearance and structure in the space of ViT features.  Our single-pair framework, which we term \emph{Splice}, does not require any additional information such as semantic segmentation and does not involve adversarial training. Furthermore, our model can be trained on high resolution images, producing high-quality results in HD. Training on a single pair allows us to deal with arbitrary scenes and objects, without the need to collect a dataset of a specific domain. We demonstrate a variety of semantic appearance transfer results across diverse natural image pairs, containing significant variations in the number of objects, pose and appearance.

While demonstrating exciting results, Splice also suffers from several limitations. First, for every input pair, it requires training a generator from scratch, which usually takes $\sim$ 20 minutes of training until convergence. This makes Splice inapplicable for real-time usage. Second, Splice is limited to observing only a single image pair and is subject to instabilities during its optimization process. Therefore, it may result in poor visual quality and incorrect semantic association in case of challenging, unaligned input pairs. To overcome these limitations, we further extend our approach to training a \emph{feed-forward} generator on a collection of \emph{domain-specific} images. Our feed-forward framework, which we term \emph{SpliceNet}, is trained directly by minimizing our novel structure and appearance ViT perceptual losses, without relying on adversarial training. SpliceNet is orders of magnitude faster than Splice, enabling real-time applications of semantic appearance transfer, and is more stable at test-time. Furthermore, due to being trained on a dataset, SpliceNet acquires better semantic association, demonstrates superior generation quality and is more robust to challenging unaligned input pairs. However, as SpliceNet is trained on a \emph{domain-specific} dataset, it is limited to working with image pairs from that domain. In contrast, Splice works with \emph{arbitrary, in-the-wild} input pairs, without any domain restriction.

We introduce two key components in the design of SpliceNet -- (i) injection of appearance information by direct conditioning on the \cls \ token feature space, and (ii) a method for distilling semantically associated structure-appearance image pairs from a diverse collection of images.

A key component in designing a feed-forward appearance transfer model is the way the network is conditioned on the input appearance image. To leverage the readily available disentangled appearance information in the \cls \ token, we design a CNN architecture that directly benefits from the information encoded in the input \cls \ token, yet controls appearance via modulation. Specifically, our model takes as input a structure image and a target \cls \ token; inspired by StyleGAN-based architectures, the content is encoded into spatial features, while the input \cls \ token is directly mapped to modulation parameters. Explicitly conditioning the model on the \cls \ token significantly simplifies the learning task, resulting in better convergence that leads to faster training and higher visual quality.

We train SpliceNet using natural image pairs from a given domain, using our DINO-ViT perceptual losses. In artistic style transfer the training examples consist of randomly sampled content and style pairs. However, in our case, the semantic association between the input images is imperative. Specifically, our training pairs should fulfill region-to-region semantic correspondence, yet differ in appearance.
Such pairs cannot be simply achieved by random pairing. Therefore, we propose an approach, leveraging  DINO-ViT features, to automatically distill such training examples out of an image collection. This allows us to train our model on diverse datasets, depicting unaligned natural poses.  We thoroughly evaluate the importance of our architectural design  and structure-appearance distillation.

\section{Related Work} \label{sec:related}

\myparagraph{Domain Transfer \& Image-to-Image Translation.} The goal of these methods is to learn a mapping between source and target \emph{domains}.
This is typically done by training a GAN on a \emph{collection} of images from the two domains,
either paired~\cite{8100115} or unpaired~\cite{8237506, 10.5555/3294771.3294838, kim2017learning, 8237572, park2020contrastive}.
Swapping Autoencoder (SA)~\cite{park2020swapping} and Kim et al. \cite{styleDiscriminator} train a domain-specific GAN to disentangle structure and texture in images,
and swap these representations between two images in the domain. These methods propose different self-supervised losses integrated in a GAN-based framework for learning disentangled latent codes from scratch.
In contrast, our method relies on disentangled descriptors derived from a pre-trained ViT feature space, and does not require any adversarial training. This significantly simplifies the learning task, allowing us to: (i) train a generator given only a single pair of images as input, while not being restricted to any particular domain, (ii) train a feed-forward model on challenging unaligned domains, in which the GAN-based methods struggle.

Recently, image-to-image translation methods trained on a single example were proposed~\cite{cohen2019bidirectional, DBLP:journals/cgf/BenaimMBW21, lin2020tuigan}. 
These methods only utilize low-level visual information and lack semantic understanding.
Our Splice framework is also trained only on a single image pair, but leverages a pre-trained ViT model to inject powerful semantic information into the generation process. Moreover, single-pair methods are based on slow optimization-processes. Our SpliceNet framework extends Splice to a feed-forward model, allowing real-time applications of semantic appearance transfer on a specific domain.

\myparagraph{Neural Style Transfer (NST).}
In its classical setting, NST transfers an \emph{artistic} style from one image to another~\cite{GatysEBHS17,JingYFYYS20}.
STROTSS~\cite{kolkin2019style} uses pre-trained VGG features to represent style and their self-similarity to capture structure in an optimization-based style transfer framework. To allow real-time use, a surge of  feed-forward models have been proposed, trained using the VGG perceptual losses  \cite{johnson2016perceptual,ulyanov2016texture,li2016precomputed,li2017diversified,dumoulin2016learned,chen2016fast,adain}. However, using second-order feature statistics results in \emph{global artistic} style transfer, and is not designed for transfering style between \emph{semantically related} regions.
In contrast, our goal is to  transfer the appearance between \emph{semantically related} objects and regions in two \emph{natural} images, which we achieve by leveraging novel perceptual losses based on a pre-trained ViT. 

\emph{Semantic style transfer} methods also aim at mapping appearance across semantically related regions between two images~\cite{MechrezTZ18, LiW16, Wilmot2017StableAC, Fast_Photographic}. 
However, these methods are usually restricted to color transformation~\cite{XuWFSZ20, Fast_Photographic, Yoo_2019_ICCV}, or depend on additional semantic inputs (e.g., annotations, segmentation, point correspondences, etc.)~\cite{GatysEBHS17, kim2020deformable, Champandard16, kolkin2019style}.  Other works tackle the problem for specific controlled domains~\cite{BarnesF14a,ShihPDF13}. In contrast, we aim to semantically transfer fine texture details in a fully automatic manner, without requiring any additional user guidance. Moreover, our Splice framework can handle arbitrary, in-the-wild input pairs, without being domain-restricted. While SpliceNet is domain-specific, it enables real-time semantic appearance transfer due to its feed-forward design.

\myparagraph{Vision Transformers (ViT).}
ViTs~ \cite{vit} have been shown to achieve competitive results to state-of-the-art CNN architectures on image classification tasks, while demonstrating impressive robustness to occlusions, perturbations and domain shifts
\cite{naseer2021intriguing}.
\dinovit~\cite{dino} is a ViT model that has been trained, without labels, using a self-distillation approach.
The effectiveness of the learned representation  has been demonstrated 
on several downstream tasks, including image retrieval and segmentation. 

\emph{Amir et al.}~\cite{amir2021deep}
have demonstrated the power of \dinovit \ Features as dense visual descriptors. Their key observation is that deep DINO-ViT features  capture rich semantic information at fine spatial granularity, e.g, describing semantic object \emph{parts}. Furthermore, they observed that the representation is shared across different yet related object classes. 
This power of \dinovit \ features was exemplified by performing ``out-of-the-box" unsupervised semantic part co-segmentation and establishing semantic correspondences across different objects categories. 
Inspired by these observations, we harness the power of \dinovit \ features in a novel generative direction -- we derive new perceptual losses capable of splicing structure and semantic appearance across semantically related objects.

\section{Method}

\begin{figure}[t!]
    \centering
    \includegraphics[width=.5\textwidth]{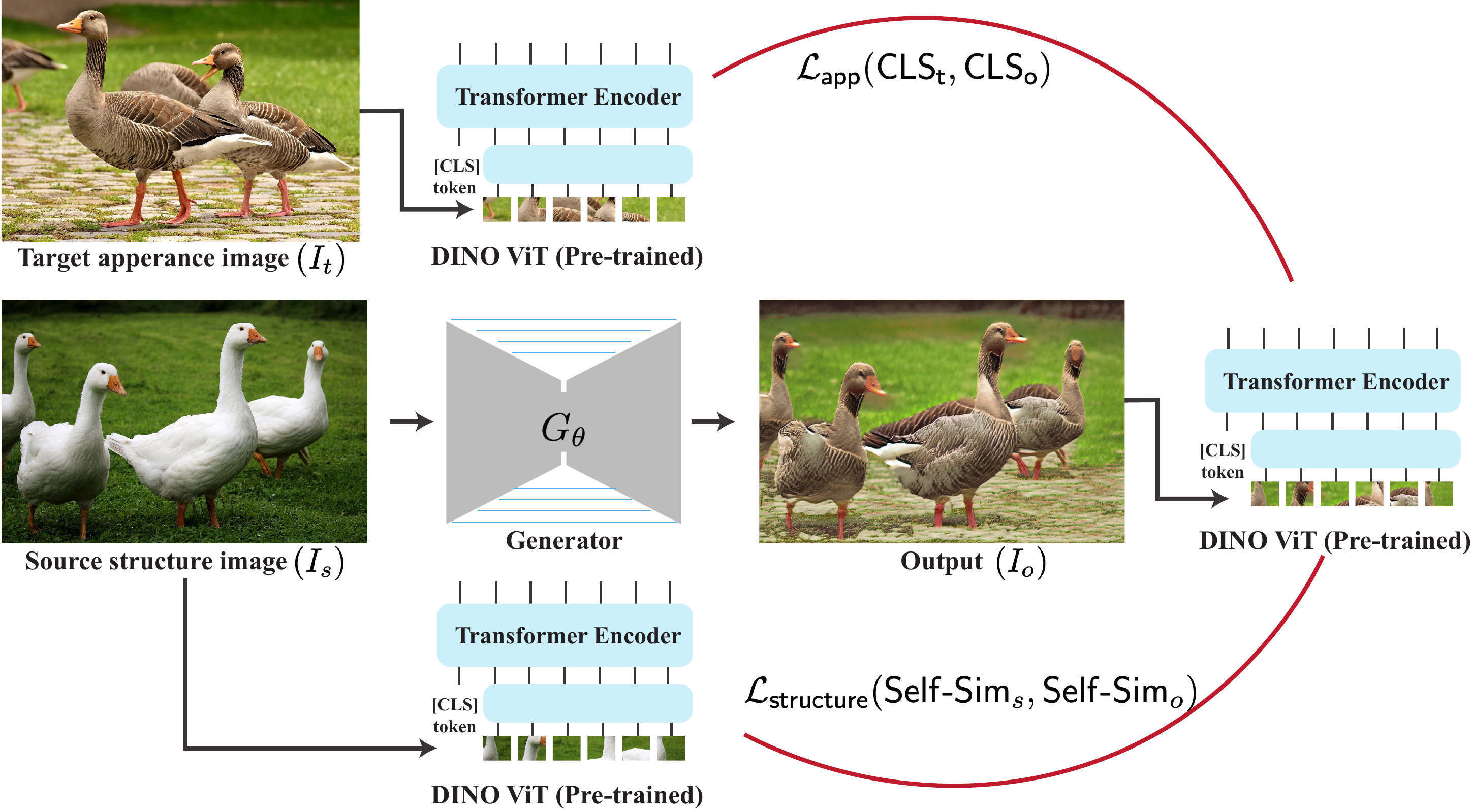}
    \caption{{\bf Splice pipeline.} Our generator $G_\theta$  takes an input structure image $I_s$ and outputs $I_o$. We establish our training losses using a pre-trained and fixed DINO-ViT model, which serves as an external semantic prior: we represent \emph{structure} via the self-similarity  of keys in the deepest attention module ($\mathsf{Self}\text{-}\mathsf{Sim}$), and \emph{appearance} via the \cls \ token in the deepest layer. Our objective is twofold: (i)   $\mathcal{L}_{\mathsf{app}}$ encourages the \cls of $I_o$ to match the \cls of $I_{t}$, and (ii) $\mathcal{L}_{\mathsf{structure}}$  encourages the self-similarity representation of $I_o$ and  $I_s$ to be the same. See Sec.~\ref{sec:method} for details.
    }
    \label{fig:pipeline} \afterfigure
\end{figure}

Given a source structure image $I_s$ and a target appearance image $I_t$, our goal is to generate an image $I_o$, in which objects in  $I_s$ are ``painted'' with the visual appearance of their semantically related objects in $I_t$. To this end, we propose Splice -- a semantic appearance transfer framework trained on a \emph{single pair} of structure and appearance images. In addition, we extend Splice to a feed-forward model trained on a dataset of images, which we term SpliceNet. While Splice can work with in-the-wild image pairs from arbitrary domains, SpliceNet is trained on a collection of images from a specific domain, and enables real-time applications due to its feed-forward design.

Our Splice framework is illustrated in Fig.~\ref{fig:pipeline}: for a given pair $\{I_s, I_t\}$, we train a generator $G_\theta(I_s)=I_o$. To establish our training losses, we leverage \dinovit~-- a  self-supervised, pre-trained ViT model~\cite{dino}~-- which is kept fixed and serves as an external high-level prior.  We propose new deep representations for \emph{structure} and \emph{appearance} in DINO-ViT feature space; we  train $G_\theta$ to output an image, that when fed into DINO-ViT, matches the source structure and target appearance representations.
Specifically, our training objective is twofold: (i) $\mathcal{L}_{\mathsf{app}}$ that encourages the deep appearance  of $I_o$ and $I_t$ to match, and (ii) $\mathcal{L}_{\mathsf{structure}}$, which encourages the deep structure  representation of $I_o$ and  $I_s$ to match.

Additionally, based on our structure and appearance losses, we design SpliceNet -- a feed-forward semantic appearance transfer framework, which is illustrated in Fig. \ref{fig:splicenet-pipeline}. The design of SpliceNet consists of two stages: a data-distillation stage, where semantically related pairs are created out of a noisy dataset, and a training stage, where we train a feed-forward generator directly conditioned on ViT feature space.

We next briefly review the ViT architecture in Sec. \ref{sec:vit}, provide qualitative analysis of \dinovit's features in Sec.~\ref{sec:inversion}, describe the Splice framework in Sec.~\ref{sec:method}, and we describe SpliceNet in Sec.~\ref{sec:splicenet_method}.

\subsection{Vision Transformers -- overview}
\label{sec:vit}

In ViT, an image $I$ is processed as a sequence of $n$ non-overlapping patches as follows: first, \emph{spatial tokens} are formed by linearly embedding each patch to a $d$-dimensional vector, and adding learned position embeddings. An additional learnable token,  a.k.a \cls \ token,  serves as a global representation of the image.

The set of tokens are then passed through $L$ Transformer layers, each consists of normalization layers (LN), Multihead Self-Attention (MSA) modules, and MLP blocks:
{\small \begin{equation*}
\begin{array}{l}
     \hat{T}^{l} = \mathsf{MSA}(\mathsf{LN}(T^{l-1}))  + T^{l-1},\\ 
     T^{l} = \mathsf{MLP}(\mathsf{LN}(\hat{T}^{l})) + \hat{T}^{l},
\end{array}
\end{equation*}} 
where $T^{l}(I)\!=\!\left[t_{\textit{cls}}^{l}(I), t_1^l(I) \dots t_n^{l}(I)\right]$ are the output tokens for layer $l$ for image $I$.

In each MSA block
the (normalized) tokens are linearly projected into queries, keys and values:
{\small \begin{equation}\label{eq:qkv}
    Q^l = T^{l-1}\cdot W_{q}^l,\ \; K^l = T^{l-1}\cdot W_{k}^l ,\ \; V^l = T^{l-1}\cdot W_{v}^l,
\end{equation}}
which are then fused using multihead self-attention to form the output of the MSA block (for full details see~\cite{vit}).

After the last layer, the \cls \ token is passed through an additional MLP to form the final output, e.g., output distribution over  a  set of labels \cite{vit}.
In our framework, we leverage \dinovit~\cite{dino}, in which the model has been trained in a self-supervised manner using a self-distillation approach. 
Generally speaking,  the model is trained to produce  the same  distribution  for  two different augmented views of the same image. As  shown in \cite{dino}, and in~\cite{amir2021deep}, \dinovit \ learns  powerful visual representations that are less noisy and more semantically meaningful than the supervised ViT.

\begin{figure*}[t!]
    \centering
    \includegraphics[width=\textwidth]{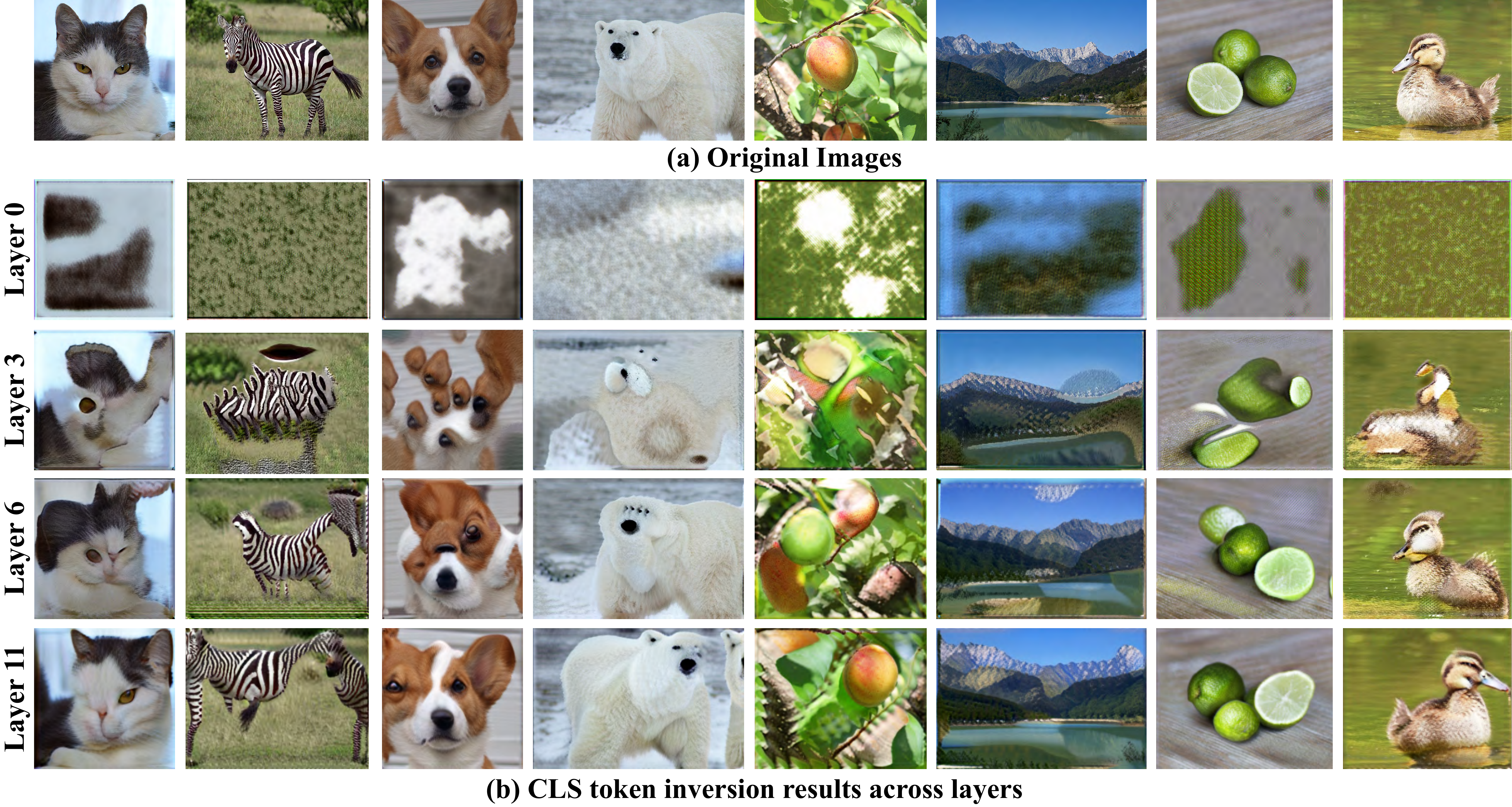}
    \caption{{\bf Inverting the [CLS] token across layers.} Each input image (a) is fed to \dinovit to compute its global [CLS] token at different layers. (b) Inversion results: starting from a noise image, we optimize for an image that would match the original [CLS] token at a specific layer. While earlier layers capture local texture, higher level information such as object parts emerges at the deeper layers (see Sec.~\ref{sec:inversion}). }\afterfigure
    \label{fig:inversion}
\end{figure*}

\subsection{Structure \& Appearance in ViT's Feature Space}
\label{sec:inversion}

The pillar of our method is the representation  of \emph{appearance} and  \emph{structure} in the space of \dinovit \ features. 
For  appearance, we want a representation that can be spatially flexible, i.e., discards the exact objects' pose and scene's spatial layout, while capturing global appearance information and style. To this end, we leverage the \cls \ token, which serves as a \emph{global} image representation.

For structure, we want a representation that is robust to local texture patterns, yet preserves the spatial layout, shape and perceived semantics of the objects and their surrounding. 
To this end, we leverage  deep \emph{spatial} features extracted from \dinovit, and use their \emph{self-similarity} as structure representation:

{\small \begin{equation}
    S^L(I)_{ij} = \mathsf{cos}\text{-}\mathsf{sim}\left( k^{L}_{i}(I), k^{L}_{j}(I)\right).  \label{eq:selfsim}
\end{equation}}
$\mathsf{cos}\text{-}\mathsf{sim}$ is the cosine similarity between keys (See Eq.~\ref{eq:qkv}).
Thus, the dimensionality of our self-similarity descriptor becomes $S^{L}(I) \in \mathbb{R}^{(n+1)\!\times\!(n+1)}$, where $n$ is the number of patches.

The effectiveness of self-similarly-based descriptors in capturing \emph{structure} while ignoring \emph{appearance} information have been previously demonstrated by both classical methods~\cite{shechtman2007localselfsim}, 
and recently also using deep CNN features for artistic style transfer~\cite{kolkin2019style}. 
We opt to use the self similarities of \emph{keys}, rather than other facets of ViT, based on~\cite{amir2021deep}. 

\begin{figure}
    \centering
    \includegraphics[width=.45\textwidth]{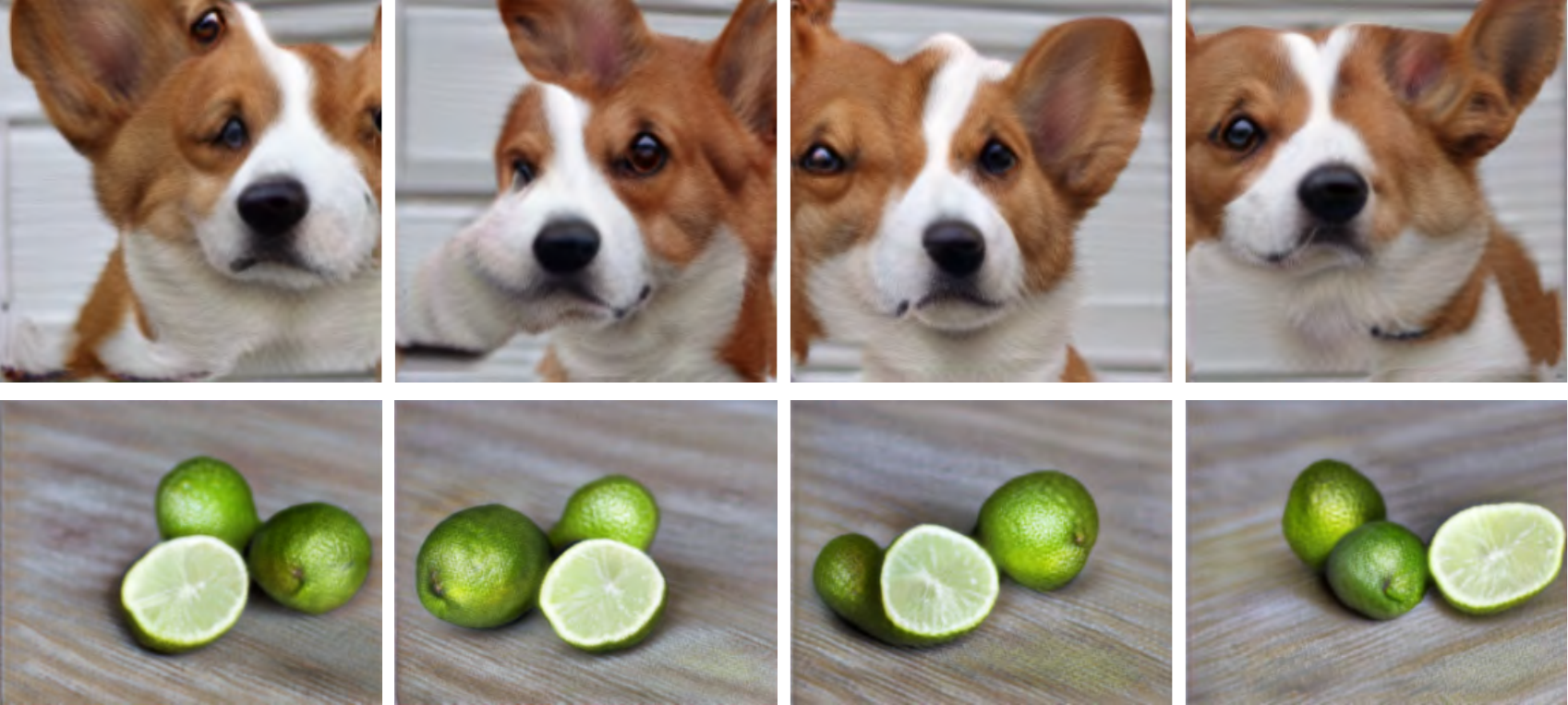}
    \caption{{\bf [CLS] token inversion over multiple runs.} The variations in structure in multiple inversion runs of the same image demonstrates the spatial flexibility of the [CLS] token.}
    \label{fig:inv_runs}\afterfigure
\end{figure}

\myparagraph{Understanding and visualizing DINO-ViT's features.} To better understand our ViT-based representations, we take a \emph{feature inversion} approach~-- given an image, we extract target features, and optimize for an image that has the same features. Feature inversion has been widely explored in the context of CNNs (e.g., \cite{simonyan2014deep,mahendran2014understanding}), however has not been attempted for understanding ViT features yet. For CNNs, it is well-known that solely optimizing the image pixels is insufficient for converging into a meaningful result~\cite{olah2017feature}. We observed a similar phenomenon when inverting ViT features (see Supplementary Materials (SM)). Hence, we incorporate ``Deep Image Prior`` \cite{UlyanovVL17}, i.e., we optimize for the weights of a CNN  $f_\theta$ that translates a fixed random noise $z$ to an output image:
{\small \begin{equation}
    \argmin_\theta ||\phi(f_\theta(z)) - \phi(I)||_F,
    \label{eq:inversion}
\end{equation}}
where $\phi(I)$ denotes the target features, and $|| \cdot ||_F$ denotes Frobenius norm.
First, we consider inverting the \cls \ token: $\phi(I)= t_{\textit{cls}}^l(I)$. Figure~\ref{fig:inversion} shows our inversion results across layers, which illustrate the following observations:
\begin{enumerate}
    \item From shallow to deep layers, the \cls \ token gradually  accumulates appearance  information. Earlier layers mostly capture local texture patterns, while in deeper layers, more global information such as object parts  emerges.
    \item The \cls \ token encodes  appearance information in a \emph{spatially flexible manner}, i.e., different object parts can stretch, deform or  be flipped.  Figure~\ref{fig:inv_runs} shows multiple runs of our inversions per image; in all runs, we can notice similar global information, but the diversity across runs demonstrates the spatial flexibility of the representation.
\end{enumerate}

Next, in Fig.~\ref{fig:keys}(a), we show the inversion of the spatial keys extracted from the last layer, i.e., $\phi(I) = K^{L}(I)$. These features have been shown to encode high level information~\cite{dino, amir2021deep}. Surprisingly, we observe that the original image can still be  reconstructed from this representation.

To discard appearance information encoded in the keys, we consider the self-similarity of the keys (see Sec.~\ref{sec:inversion}). 
This is demonstrated in the PCA visualization of the keys' self-similarity in Fig.~\ref{fig:keys}(b). As seen, the self-similarity mostly captures  the  structure of objects, as well as their distinct semantic components. For example, the legs and the body of the polar bear that have the same texture, are distinctive.

\subsection{Splicing ViT Features}
\label{sec:method}

Based on our understanding of \dinovit's internal representations, we turn to the task of training a generator given a single pair of structure-appearance images. Our framework, which we term Splice, is illustrated in Fig. \ref{fig:pipeline}.

Our objective function takes the following form:

{\small \begin{equation} \label{eq:4}
    \mathcal{L}_{\mathsf{splice}} =   \mathcal{L}_{\mathsf{app}} + \alpha \mathcal{L}_{\mathsf{structure}} + \beta \mathcal{L}_{\mathsf{id}},
\end{equation}}
where $\alpha$ and $\beta$ set the relative weights between the terms. We set $\alpha=0.1, \beta=0.1$ for all experiments of Splice.

\myparagraph{Appearance loss.} The term $\mathcal{L}_{\mathsf{app.}}$ encourages the output image to match the appearance of $I_t$, and is defined as the difference in \cls  token between the generated and appearance image: 
{\small
\begin{equation} \label{eq:6}
    \mathcal{L}_{\mathsf{app}} = \left\| t_{\mathcls}^{L}(I_t) - t_{\mathcls}^{L}(I_o) \right\|_2 ,
\end{equation}}
where $t_{\mathcls}^{L}(\cdot) = t_{cls}^{L}$ is the \cls \ token extracted from the deepest layer (see Sec.~\ref{sec:vit}).

\myparagraph{Structure loss.} The term $\mathcal{L}_{\mathsf{structure}}$ encourages the output image to match the structure of $I_s$, and is defined by the difference in self-similarity of the keys extracted from the attention module at deepest transformer layer: 

\begin{figure}[t!]
    \centering
    \includegraphics[width=.5\textwidth]{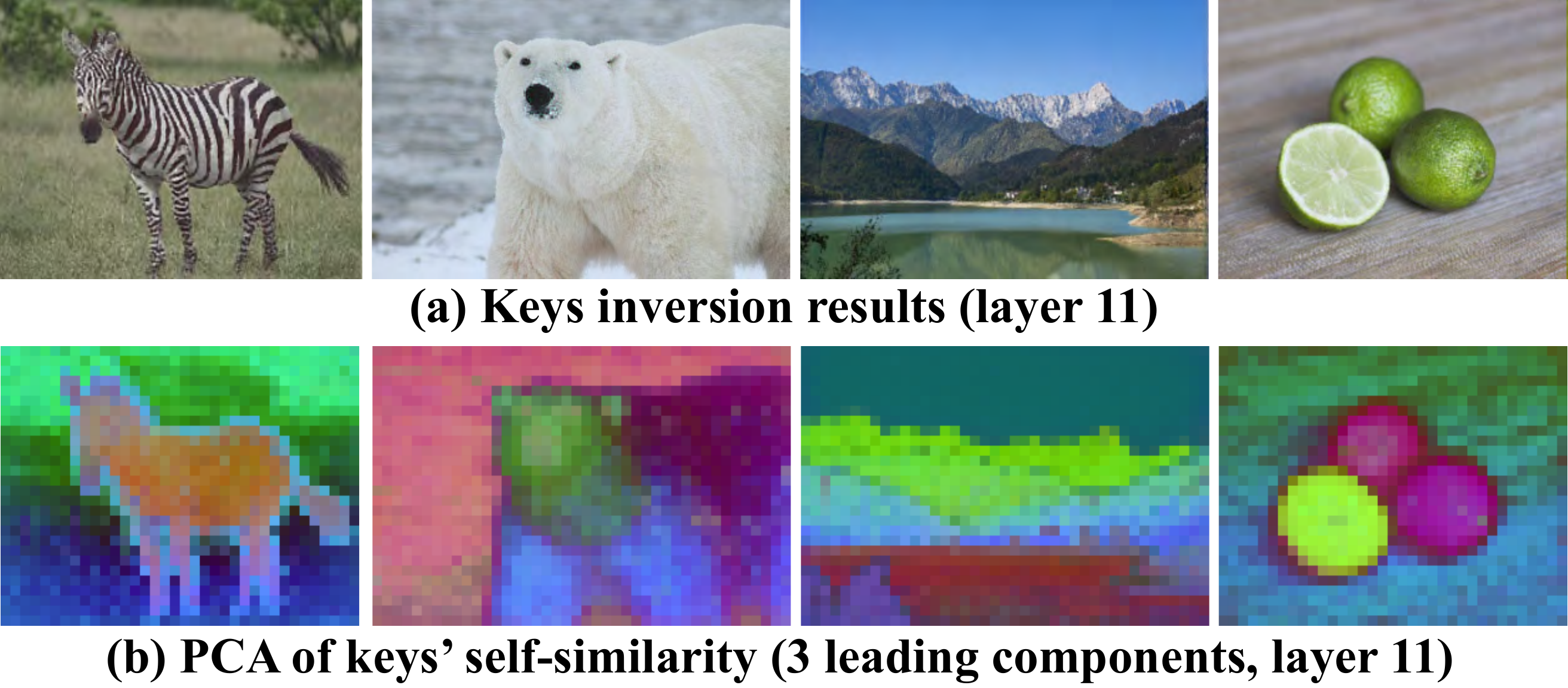}
    \caption{{\bf Visualization of DINO-ViT keys.} (a) Inverting keys from the deepest layer surprisingly reveals  that the  image can be reconstructed.  (b) PCA visualization of the  keys' self-similarity: the leading components mostly  capture semantic scene/objects parts, while discarding appearance information (e.g., zebra stripes). } \afterfigure
    \label{fig:keys}
\end{figure}

\begin{figure*}[t!]
    \centering
    \includegraphics[width=1\textwidth]{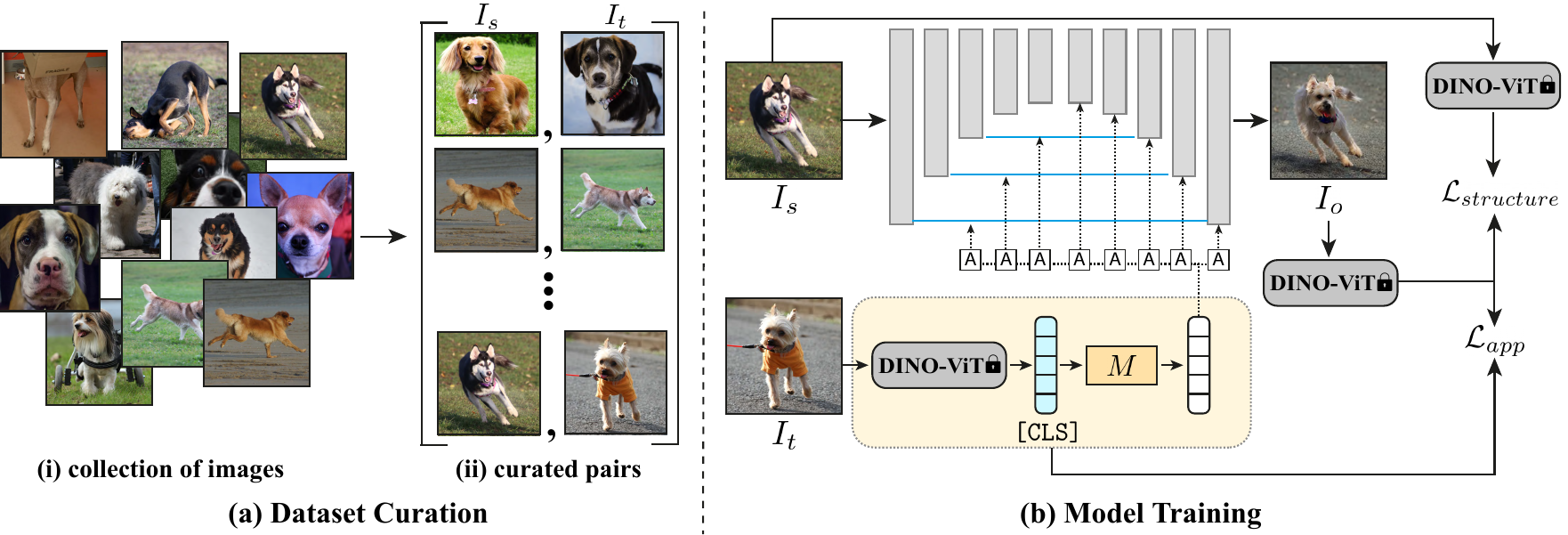} \vspace{-0.5cm}
    \caption{SpliceNet Pipeline. (a) A diverse image collection is automatically curated for distilling image pairs used for training, each depicts region-to-region semantic correspondences as well as significant variation in appearance. (b) SpliceNet comprises of a  UNet architecture, which takes as input: a structure image ($I_s$), and the \cls \ token extracted from a pre-trained DINO-ViT when fed with the target appearance image ($I_t$). The structure image is  encoded into spatial features, while the \cls \ token is used to adaptively normalize the decoded features. This is done via a mapping network ($M$) followed by learnable affine transformations~\cite{stylegan2}. Skip connections are used to allow the model to retain fine content details. Our model is trained using DINO-ViT perceptual losses: (1) $\mathcal{L}_{\mathsf{app}}$ that encourages the appearance of $I_o$ and $I_t$ to match, and (2) $\mathcal{L}_{\mathsf{structure}}$, which encourages the structure and perceived semantics of $I_o$ and  $I_s$ to match. See Sec. \ref{sec:method} for details.
    }\afterfigure
    \label{fig:splicenet-pipeline}

\end{figure*}

\begin{figure}  
    \centering
    \includegraphics[width=0.45\textwidth]{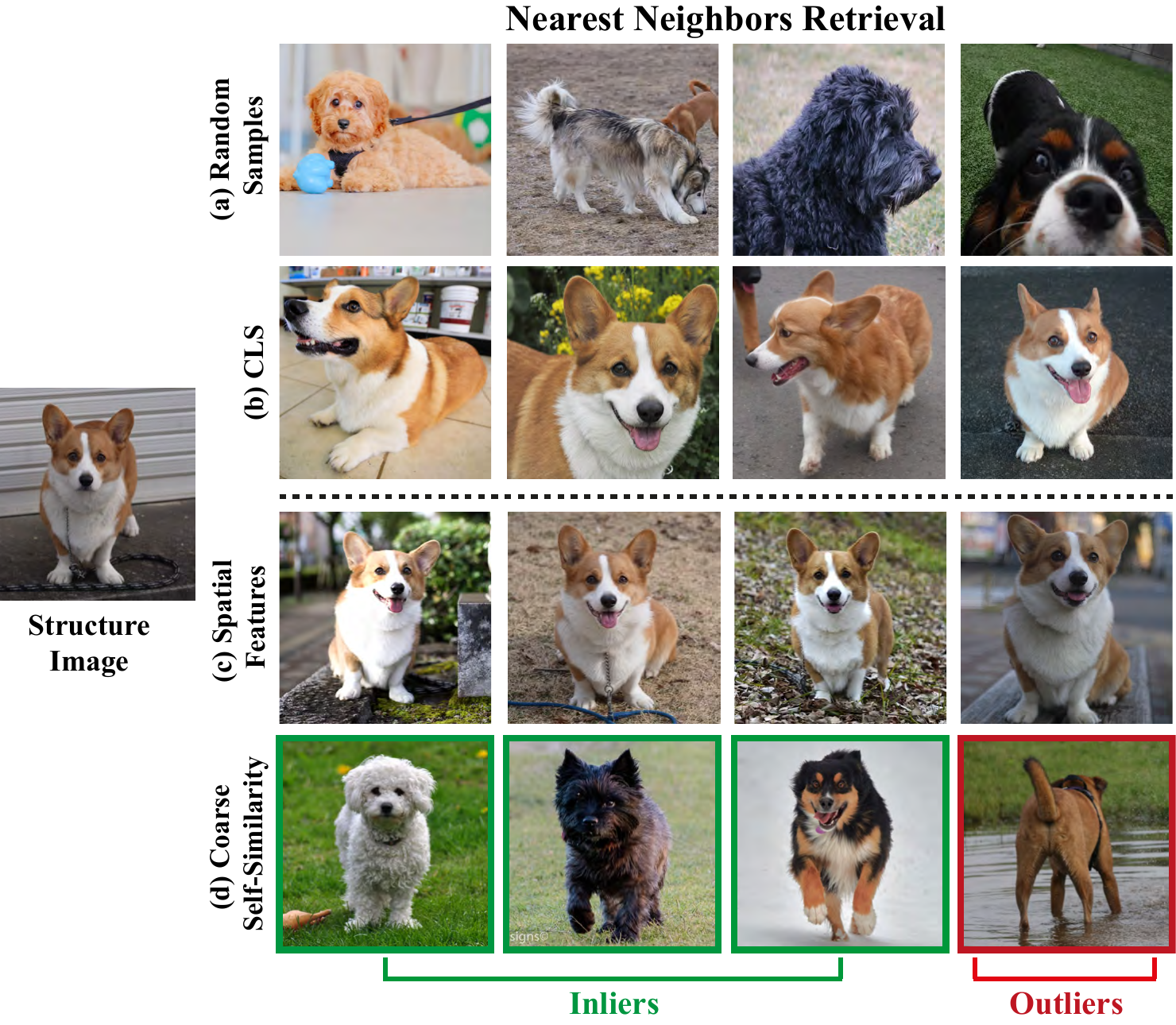}\vspace{-0.3cm}
    \caption{We retrieve K-nearest-neighbors of a structure image (left) using similarity in (b) \cls \ token. (c) spatial features. (d) Coarse Self-Similarity. Inlier  are marked in green, whereas outliers in red. See Sec. \ref{sec:data} for details.} \afterfigure
    \label{fig:pairing}
\end{figure}

{\small \begin{equation} \label{eq:5}
    \mathcal{L}_{\mathsf{structure}} = \left\|S^{L}(I_s) - S^{L}(I_o) \right\|_F,
\end{equation}}
where $S^{L}(I)$ is defined in Eq.~(\ref{eq:selfsim}).
\begin{figure*}
    \centering
    \includegraphics[width=\textwidth]{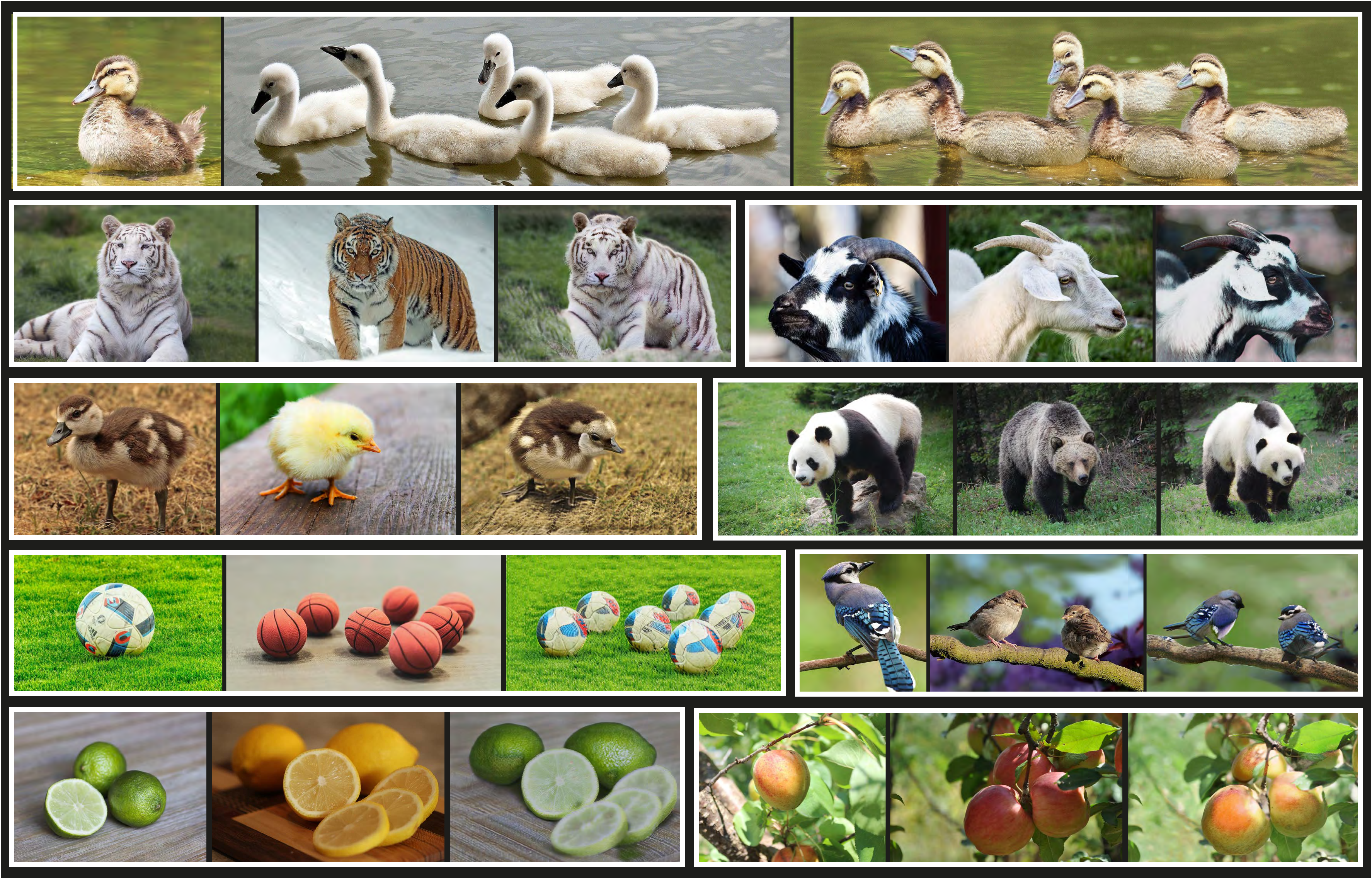}
    \caption{{\bf Sample results of Splice on in-the-wild image pairs.} For each example, shown left-to-right: the target appearance image, the source structure image and our result. The full set of results is included in the SM. Notice the variability in number of objects, pose, and the significant appearance changes between the images in each pair. } \afterfigure 
    \label{fig:results}
\end{figure*}

\begin{figure*}[t!]
    \centering
    \includegraphics[width=\textwidth]{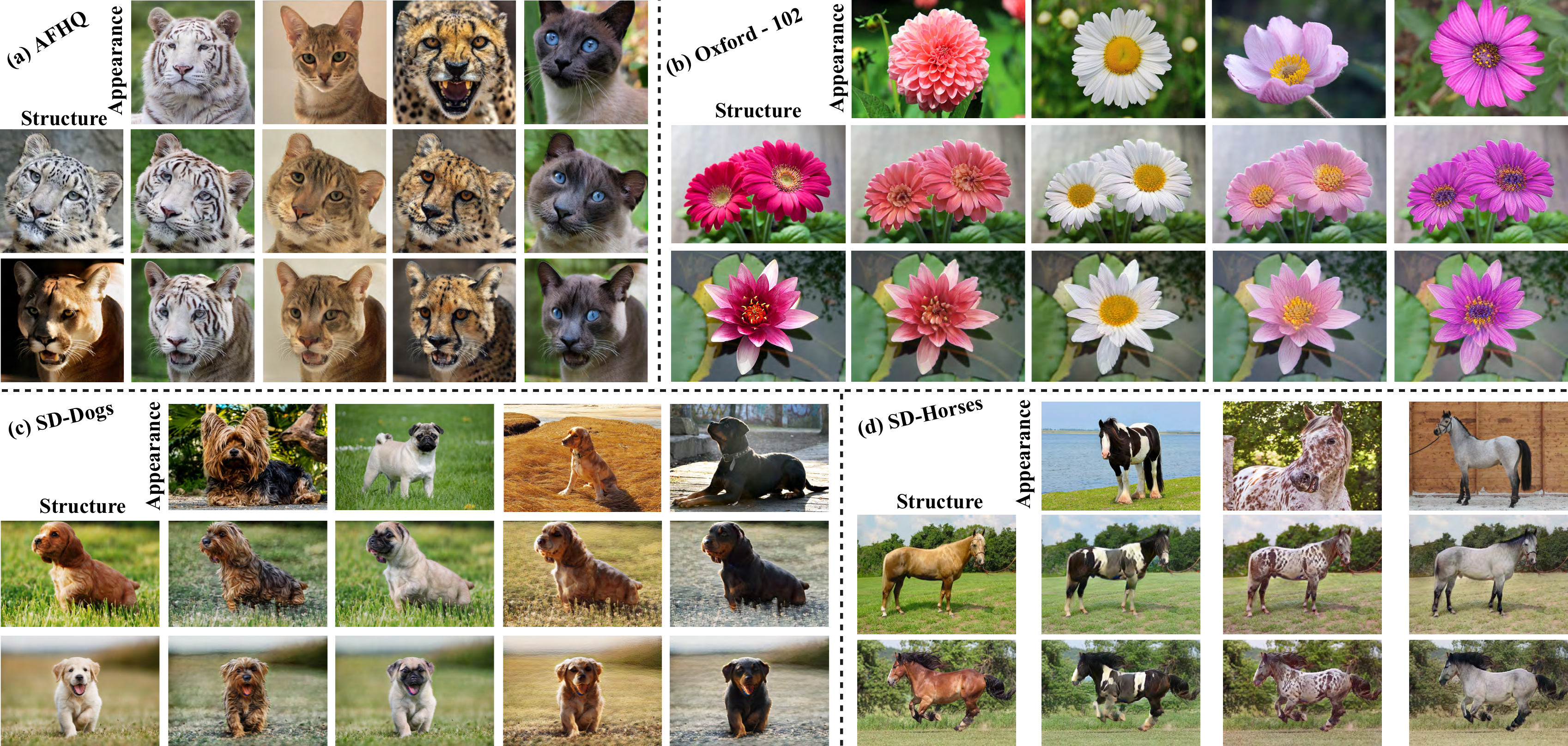}\vspace{-0.2cm}
    \caption{Sample results of SpliceNet trained for (a) \emph{AFHQ}, (b) \emph{Oxford-102} (c) \emph{SD-Dogs} and (d) \emph{SD-Horses}. Across rows: different \emph{structure} images, across columns: different \emph{appearances}. The full set of results is included in the SM.} \afterfigure
    \label{fig:splicenet-results}
\end{figure*}

\myparagraph{Identity Loss.} The term $\mathcal{L}_{\mathsf{id}}$ is used as a regularization. Specifically, when we feed $I_t$ to the generator, this loss encourages $G_\theta$ to preserve the keys representation of $I_t$: 
{\small \begin{equation} \label{eq:7}
    \mathcal{L}_{{\mathsf{id}}} = \left\| K^{L}(I_t) - K^{L}(G_\theta(I_t)) \right\|_F. \\
\end{equation}}

Similar loss terms, defined in RGB space, have been used as a regularization in training GAN-based generators for image-to-image translation \cite{park2020contrastive, DBLP:conf/iclr/TaigmanPW17, 8237506}. Here, we apply the identity loss with respect to the \emph{keys} in the deepest ViT layer, a semantic yet invertible representation of the input image (as discussed in section \ref{sec:inversion}).

\myparagraph{Data augmentations and training.} Since we only have a single input pair $\{I_s, I_t\}$, we create additional training examples, $\{I^i_s, I^i_t\}_{i=1}^N$, by applying augmentations such as crops and color jittering (see Appendix \ref{sec:appendix-data-aug} for implementation details).  $G_\theta$ is now trained  on multiple \emph{internal examples}. Thus, it has to learn a good mapping function for a \emph{dataset} containing $N$ examples, rather than solving a test-time optimization problem for a single instance.  Specifically, for each example, the objective is to generate $I_o^i = G_\theta(I_s^i)$, that matches the structure of $I_s^i$ and the appearance of $I_t^i$.

\subsection{SpliceNet: A Feed-forward Model for Semantic Appearance Transfer}
\label{sec:splicenet_method}

While Splice demonstrates exciting results on in-the-wild image pairs as shown in Sec. \ref{sec:results} and Figures \ref{fig:teaser},\ref{fig:results},\ref{fig:comparisons}, it requires training a generator from scratch for each structure-appearance image pair. This costly optimization process makes the framework infeasible for real-time applications. To this end, we propose SpliceNet -- a feed-forward appearance transfer framework. SpliceNet is a feed-forward generator trained on a dataset of images with diverse alignment and appearance, and its objective function is based on the perceptual losses described in Sec. \ref{sec:method}. While being domain-specific, SpliceNet is orders of magnitude faster than Splice at inference time, allowing real-time applications of semantic appearance transfer.

Given a source structure image $I_s$ and a target appearance image $I_t$ in domain $\mathcal{X}$, we seek a feed-forward model $F_\theta$ that outputs a stylized image~${I_o}$. 
A straightforward approach is to directly condition $F_\theta$ on the input source-target images themselves, i.e., $I_o = F_\theta(I_s; I_t)$. However, the model would have to implicitly learn to extract appearance information from $I_t$, while discarding irrelevant spatial information -- a challenging task by itself. Instead, our key observation is that such a representation is readily available in DINO-ViT's \cls \ token, which can serve as an input to the model, i.e., $I_o = F_\theta\left(I_s; t_{\mathcls}^{L}(I_t)\right)$. Directly conditioning the model on the \cls \ token significantly simplifies the learning task, resulting in better convergence that leads to faster training and higher visual quality. We thoroughly analyze the effectiveness of this design in Sec.~\ref{sec:ablation}. 

Specifically, our framework, illustrated in Fig.~\ref{fig:splicenet-pipeline}, consists of a U-Net architecture \cite{ronneberger2015u}, which takes as input the structure image $I_s$, and a \cls \ token $t_{\mathcls}^{L}(I_t)$. The structure image is encoded and then decoded to the output image, while the \cls \ token is used to modulate the decoder's feature. This is done by feeding $t_{\mathcls}^{L}(I_t)$ to a 2-layer MLP ($M$) followed by learnable affine transformations~\cite{stylegan2}. See more details in Appendix \ref{sec:appendix-splicenet-architecture}.

\subsection{Structure-Appearance Pairs Distillation \& Training}
\label{sec:data}
An important aspect in training our model to transfer appearance across natural images is \emph{data}. While diverse natural image collections are available, 
randomly sampling structure-appearance image pairs and using them as training examples is insufficient. Such random pairs often cannot be semantically associated (e.g., a zoomed-in face of a dog vs. a full-body, as seen Fig.~\ref{fig:pairing} top row). Thus, training a model with high prevalence of such pairs prevents it from learning meaningful semantic association between the structure and appearance images. We tackle this challenge by \emph{automatically} distilling image pairs $(I_s, I_t)$ that satisfy the following criteria: (i)~depict semantic region-to-region correspondence, and (ii)~substantially differ in appearance, to encourage the network to utilize the rich information encoded by \cls \ token, and learn to synthesize complex textures.

To meet the above criteria, we need an image descriptor $\mathcal{F}(I)$, invariant to appearance, that can capture the rough semantic layout of the scene. To this end, we leverage the DINO-ViT representation, and use a spatially-coarse version of keys' self-similarity as image descriptor. That is,
\begin{equation}
    \mathcal{F}(I) = S_{\text{coarse}}(I) \in \mathbb{R}^{d\times\!d},
\end{equation}
where $S_{\text{coarse}}(I)$ is the self-similarity matrix computed by average pooling the grid of spatial keys, and then plugging the pooled keys, $\bar{K}^L(I)$, to Eq.~\ref{eq:selfsim}; here $d=\sqrt{n} / w$, where $n$ is the number of spatial features, and $w$ is the pooling window size. 

Figure~\ref{fig:pairing} shows top-4 nearest-neighbors retrieved using different descriptors for a given query image. As seen in Fig.~\ref{fig:pairing}(c), directly comparing the features results in similar semantic layout, yet all the images depict very similar appearance. Using  coarse self-similarity, we obtain a set of images spanning diverse appearances. Furthermore, using \emph{coarse} feature map allows for more variability in the pose of the dogs, which further increases the diversity of our pairs (Fig.~\ref{fig:pairing}(d)).

A simple structure-appearance pairing could be achieved by pairing each image $I\in \mathcal{X}$ with its K-nearest-neighbors (KNN) according to the similarity in $\mathcal{F}(I)$. However, such approach does not account for outlier images, which often appear in Internet datasets. To this end,  we use a robust similarity metric based on the \emph{Best-Buddies Similarity} (BBS)~\cite{dekel2015best}, in which an image pair $(I_i, I_j)$ is considered as inlier if the two images are mutual nearest neighbours. Here, we extend this definition to mutual $K$- nearest-neighbors, and pair each query image $I_q$ with a set of images $\{I_j\}$ that satisfy:
\begin{equation}
 I_j \in KNN(I_q, \mathcal{X})~~\land~~I_q \in KNN(I_j, \mathcal{X}).
\end{equation}
Fig. \ref{fig:pairing} (bottom row) shows an example of automatically detected inliers/outliers. More examples are included in the SM. 

\paragraph{Training.} At each training step, we sample an image pair $(I_s, I_t)$ from our distilled paired-dataset and apply various augmentations, such as cropping and flipping (see Appendix \ref{sec:appendix-data-aug} for full details). We then feed $I_t$ to DINO-ViT and extract the \cls \ token $t_{\mathcls}^{L}(I_t)$, which is fed to our model $I_o = F_\theta\left(I_s; t_{\mathcls}^{L}(I_t)\right)$; our training objective is given in Eq.~\ref{eq:4}. $\mathcal{L}_{\mathsf{structure}}$ and $\mathcal{L}_{\mathsf{app}}$ have the same definitions as in Splice. $\mathcal{L}_{\mathsf{id}}$ is used as a reconstruction regularization term in case $I_s = I_t = I$, i.e.,  $\mathcal{L}_{\mathsf{identity}} = \mathcal{D}\left(I, I_o\right)$, where $\mathcal{D}(\cdot,\cdot)$ is a chosen distance function.  Empirically we found LPIPS~\cite{lpips} to be more stable compared to the keys loss described in Sec. \ref{sec:method}.

\section{Results}
\label{sec:results}

\vspace{0.1cm}

\subsection{Splice}
\myparagraph{Datasets.} We tested Splice on a variety of image pairs gathered from Animal Faces
HQ (AFHQ) dataset \cite{Choi_2020_CVPR}, and images crawled from \href{https://flickr.com/groups/justmountains/pool/}{Flickr Mountain}. In addition, we collected our own dataset, named \emph{Wild-Pairs}, which includes a set of 25 high resolution image pairs taken from \href{https://pixabay.com}{Pixabay}, each pair depicts semantically related objects from different categories including animals, fruits, and other objects.  The number of objects, pose and appearance may significantly change between the images in each pair.  The image resolution ranges from 512px to 2000px.

Sample pairs from our dataset along with our results can be seen in Fig.~\ref{fig:teaser} and Fig.~\ref{fig:results}, and the full set of pairs and results is included in the SM. As can be seen, in all examples, our method  successfully transfers the visual appearance in a semantically meaningful manner at several levels:  (i) \emph{across objects:} the target visual appearance of objects is being transferred to to their semantically related objects in the source structure image, under significant variations in pose, number of objects, and appearance between the input images.  (ii) \emph{within objects:} visual appearance is transferred between corresponding body parts or object elements. For example, in Fig.~\ref{fig:results} top row, we can see the appearance of a single duck is semantically transferred to each of the 5 ducks in the source image, and that the appearance of each body part is mapped to its corresponding part in the output image.  This can be consistently observed in all our results. 

The results  demonstrate that our method is capable of performing semantic appearance transfer across diverse image pairs, unlike GAN-based methods which are restricted to the dataset they have been trained on.

\subsection{SpliceNet}

\myparagraph{Datasets.} We trained SpliceNet on the training set of each of the following (separately): Animal Faces HQ (\emph{AFHQ}) \cite{Choi_2020_CVPR}, \emph{Oxford-102} \cite{oxford102}, and two Internet datasets~-- \emph{SD-Dogs}, and \emph{SD-Horses}~-- each containing a wide range of poses,  and appearance variations~\cite{mokady2022selfdistilled}. 

Fig.~\ref{fig:results} shows sample results of our method on diverse structure-appearance pairs.  SpliceNet consistently transfers appearance between semantically-corresponding regions, while synthesizing high-quality textures. Notably, although the appearance may dramatically change, the structure and perceived semantics of the content image are well preserved across all datasets. Many more results are included in SM.

\subsection{Comparisons} \label{sec:comparison}

For \textbf{Splice}, there are no existing methods that are tailored for solving its task:  semantic appearance transfer between two natural images (not restricted to a specific domain), without explicit user-guided inputs. We thus compare Splice to prior works in which the problem setting is most similar to ours in some aspects (see discussion in these methods in Sec.~\ref{sec:related}):
(i) \emph{Swapping Autoencoders (SA)}~\cite{park2020swapping} -- a domain-specific, GAN-based method which has been trained to ``swap'' the texture and structure of two images in a realistic manner; (ii)  \emph{STROTSS}~\cite{kolkin2019style}, the style transfer method that also uses self-similarity of a pre-trained CNN features as the content descriptor, (ii) \emph{WCT$^2$}~\cite{Yoo_2019_ICCV}, a photorealistic NST method.

Since SA requires a dataset of images from two domains to train, we can only compare our results to their trained models on AHFQ and Flicker Mountain datasets.  For the rest of the methods, we also later compare to image pairs from our \emph{Wild-Pairs} examples. We evaluate our performance across a variety of image pairs both qualitatively, quantitatively and via an AMT user study. 

We compare \textbf{SpliceNet} to prior works in which the problem setting is most similar: Splice; WCT$^2$~\cite{Yoo_2019_ICCV}; and prominent GAN-based methods: Swapping Autoencoder (SA)~\cite{park2020contrastive} and \cite{styleDiscriminator}. We used official implementation and pre-trained models of these methods when available. We trained SA and \citeauthor{styleDiscriminator} for the datasets for which no model was provided by the authors. 

Table~\ref{table:reconstruction}(left) reports the number of trainable parameters in each of these models, and their average inference run-time.

\subsubsection{Qualitative comparison}

\begin{figure*} [h!]
    \centering
    \includegraphics[width=1\textwidth]{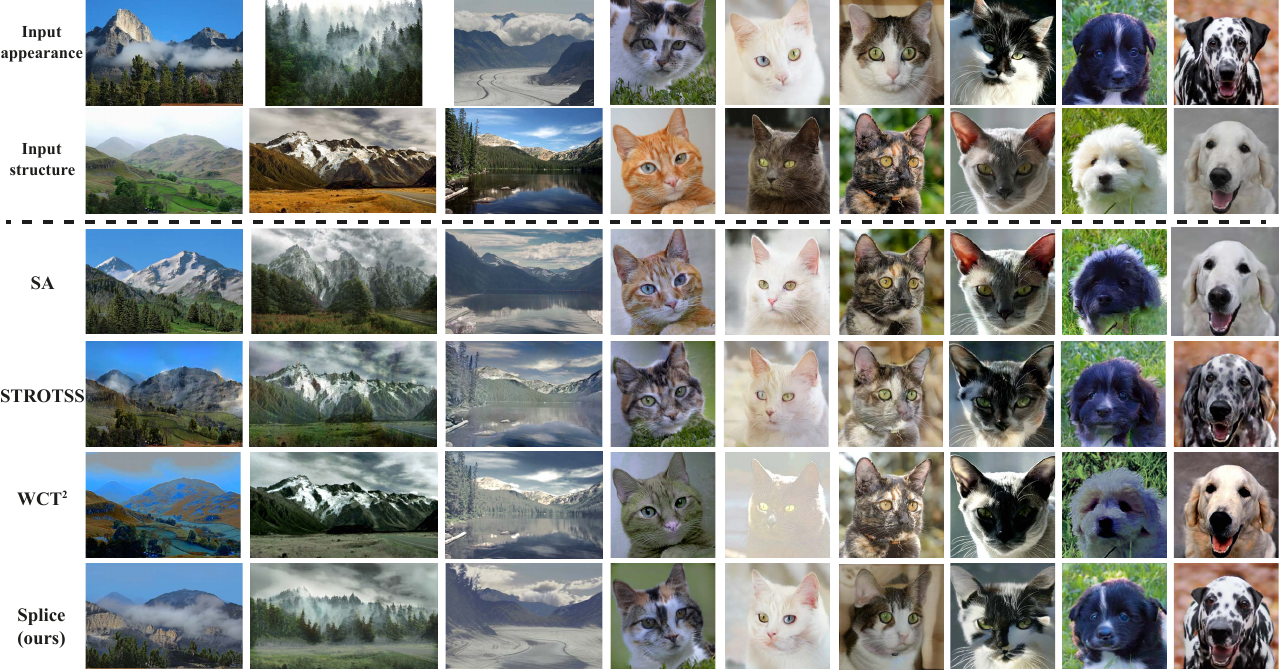}
    \caption{{\bf Comparisons of Splice with style transfer and swapping autoencoders.} First two rows: input appearance and structure images taken from the AFHQ and Flickr Mountains. The following rows, from top to bottom, show the results of: swapping autoencoders (SA) \cite{park2020swapping},  STROTSS \cite{kolkin2019style}, and  $\text{WCT}^2$ \cite{Yoo_2019_ICCV}.  See SM for additional comparisons. } \afterfigure

    \label{fig:comparisons}
\end{figure*}

Figure~\ref{fig:comparisons} shows sample results for all methods (additional results are included in the SM) compared to \textbf{Splice}. In all examples, Splice correctly relates semantically matching regions between the input images, and successfully transfers the visual appearance between them.  In the landscapes results (first 3 columns), it can be seen that SA outputs high quality images but  sometimes struggles to maintain high fidelity to the structure and appearance image: elements for the appearance image are often missing e.g., the fog in the left most example, or the trees in the second from left example. These visual elements are captured well in our results. For AHFQ, we noticed that SA often outputs a result that is nearly identical to the structure image. A possible cause to such behavior might be the adversarial loss, which ensures that the swapping result is a realistic image according to the the distribution of the training data. However, in some cases, this requirement does not hold (e.g. a German Shepherd with leopard's texture), and by outputting  the structure image the adversarial loss can be trivially satisfied.\footnote{We verified these results with the  authors \cite{park2020swapping}}.

NST frameworks such as STROTSS and WCT$^2$ well preserve the structure of the source image, but their results often depict visual artifacts:  STROTSS's results often suffer from color bleeding artifacts, while  WCT$^2$ results in  global color artifacts, demonstrating that transferring color is insufficient for tackling our task. 

Splice demonstrates better fidelity to the input structure and appearance images than GAN-based SA, while training only on the single input pair, without requiring a large collection of examples from each domain. With respect to style transfer, Splice better transfers the appearance across semantically related regions in the input images, such as matching facial regions (e.g., eyes-to-eyes, nose-to-nose), while persevering the source structure. 

\begin{figure} [t!]
    \centering
    \vspace{0.3cm}
    \includegraphics[width=.4\textwidth]{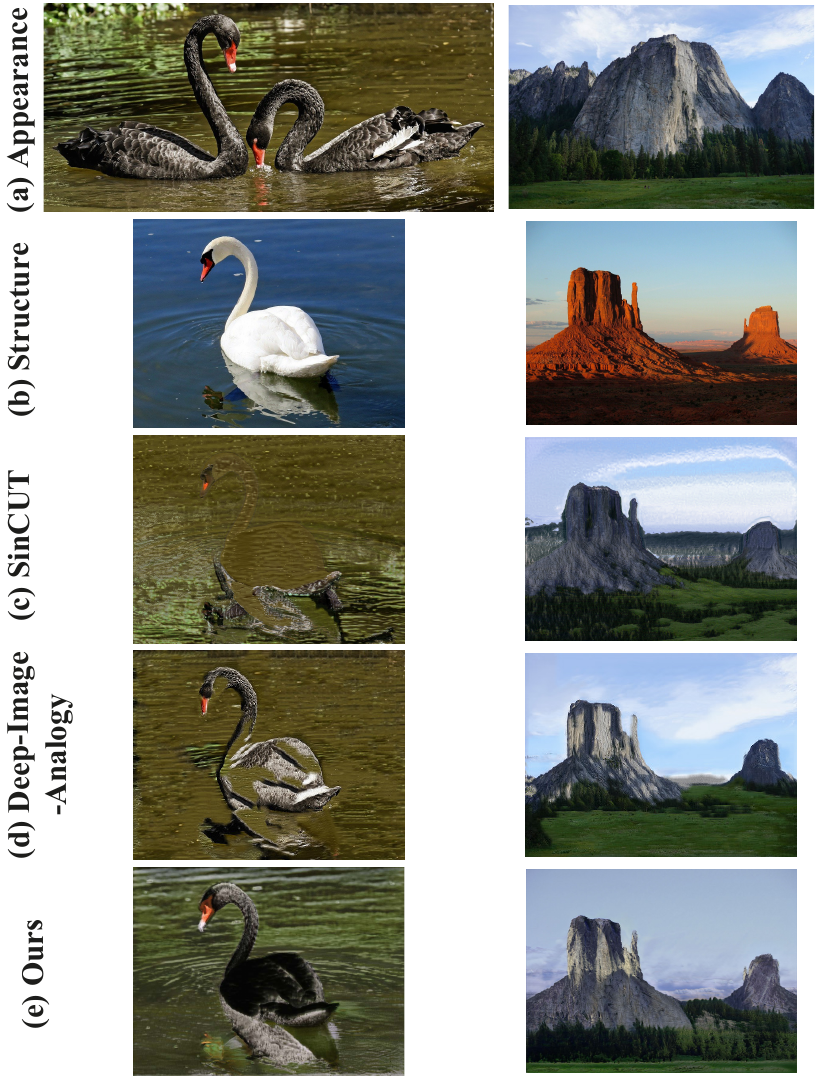} \vspace{-0.3cm}
    \caption{{\bf Additional Qualitative Comparisons.}  SinCUT ~\cite{park2020contrastive} (c) and Deep-Image-Analogy ~\cite{Liao2017} (d) results, when trained on each input pair (a-b). These methods work well when the translation is mostly based on low-level information (top), but fail when higher-level reasoning is required (bottom), struggling to make meaningful semantic associations (e.g., the lake is mapped to the swan). (e) Our method successfully transfers the appearance across semantic regions, and generates high-quality results w/o adversarial training.}\afterfigure
    \label{fig:sincut_comp}
\end{figure}

Finally, we also include qualitative comparisons to SinCUT~\cite{park2020contrastive}, a GAN-based image translation method, and to Deep-Image-Analogy ~\cite{Liao2017}. As demonstrated in Fig.~\ref{fig:sincut_comp},  SinCUT and Deep-Image-Analogy perform well for the landscape example,

but fail to transfer the appearance of the swan in the second example, where a higher-level visual understanding is required. Splice successfully transfers the appearance across semantically realted regions, and generates high quality results w/o adversarial loss.

\begin{figure*}[t!]
    \centering
    \includegraphics[width=.9\textwidth]{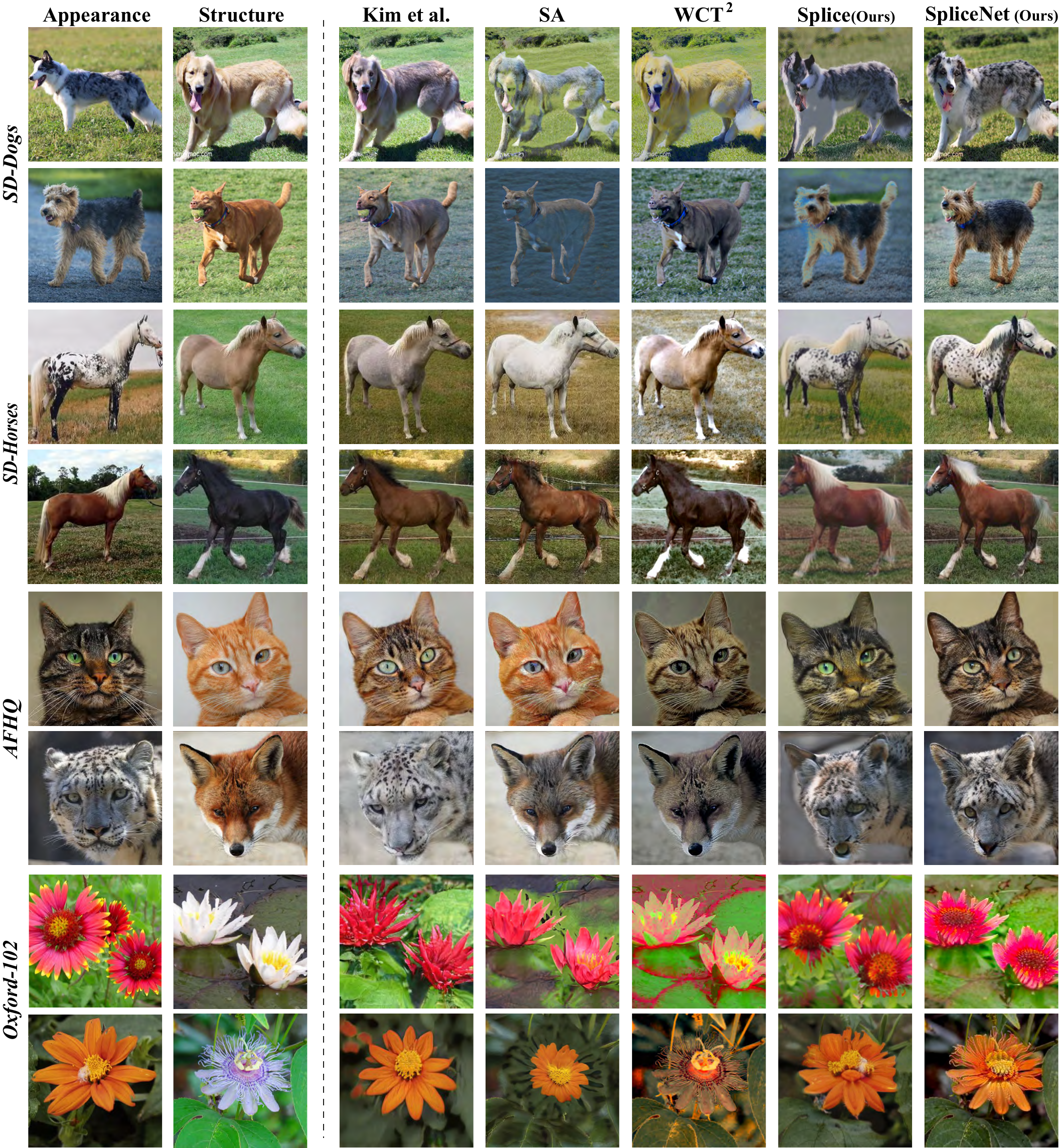}\vspace{-0.25cm}
    \caption{{\bf SpliceNet comparisons with baselines.} First two columns depict the input appearance-structure. The following columns show the results of: \cite{styleDiscriminator}, Swapping Autoencoder (SA) \cite{park2020swapping}, WCT$^2$ \cite{Yoo_2019_ICCV}, Splice, and SpliceNet. Top to bottom: \emph{SD-Dogs}, \emph{SD-Horses}, \emph{AFHQ} and \emph{Oxford-102}. See SM for additional comparisons.} \vspace{-0.3cm}
    \label{fig:splicenet-comparisons}
\end{figure*}


Figure~\ref{fig:comparisons} shows comparison between \textbf{SpliceNet} and baselines. 
As seen by WCT$^2$ results, transferring colors is insufficient for capturing the target appearance. The GAN-based methods (SA and Kim et.~al.), which learn structure/appearance representations from scratch, suffer from either bleeding artifacts or low fidelity to the source structure for aligned datasets (\emph{AFHQ}, and \emph{Oxford-102}). For more diverse and unaligned datasets (\emph{SD-Dogs} and \emph{SD-Horses}), these methods struggle to synthesize complex textures or to preserve the original content. 
Although Splice can successfully establish semantic association, it is subject to instabilities in its test-time optimization process that sometimes leads to failure cases (e.g., topmost flower, second dog), while SpliceNet achieves improved visual quality and stability.

\subsubsection{Quantitative comparison}
\label{sec:quantitive}

To quantify how well our generated images match the target appearance and  preserve the original structure, we use the following metrics: (i) human perceptual evaluation, (ii) semantic layout preservation and (iii) reconstruction.

\myparagraph{Human Perceptual Evaluation} We design a user survey suitable for evaluating the task of  appearance transfer across semantically related scenes. We adopt the Two-alternative Forced Choice (2AFC) protocol suggested in \cite{park2020swapping, kolkin2019style}. Participants are shown with 2 reference images: the input structure image (A), shown in grayscale, and the input appearance image (B), along with 2 alternatives: our result and another baseline result. The participants are asked: \emph{``Which image best shows the shape/structure of~image~A combined with the appearance/style of image~B?''}.

For evaluating \textbf{Splice}, we perform the survey using a collection of 65 images in total, gathered from AFHQ, Mountains, and Wild-Pairs. We collected 7000 user judgments w.r.t. existing baselines. Table \ref{table:amt} reports the percentage of votes in our favor.  As seen, our method outperforms all baselines across all image collections, especially in the Wild-Pairs, which highlights our performance in challenging settings. Note that SA was trained on 500K mountain images, yet our method perform competitively.

{\small \begin{table*}[t!] 
\centering
\setlength{\tabcolsep}{4pt}
\begin{tabular}{c|c|c||c|c|c|c|c|c|c|c|c|c|c|c} 
 \hline
 & Params& Runtime & \multicolumn{3}{c|}{\emph{AFHQ}} & \multicolumn{3}{c|}{\emph{Oxford-102}} & \multicolumn{3}{c|}{\emph{SD-Horses}} & \multicolumn{3}{c}{\emph{SD-Dogs}} \\ 
\cline{4-15}
 & [M] & [sec] & MSE & LPIPS & Human eval & MSE & LPIPS & Human eval & MSE & LPIPS & Human eval & MSE & LPIPS & Human eval \\
 \hline
 \citeauthor{styleDiscriminator} & 56.51 
 & .1251 & .0506 & .2053 & 86.79 $\pm$ 0.23 & .0817 & .3658 & 80 $\pm$ 0.17 & .0251 & .1350 & 73.84 $\pm$ 0.16 & .0276 & .1707 & 91.1 $\pm$ 0.2 \\
 SA & 109.03 
 & \underline{.0954} & .0241 & .1452 & 98.93 $\pm$ 0.03 & .0355 & .1745 & 90.29 $\pm$ 0.21 & .0454 & .2464 & 96.53 $\pm$ 0.08 & .0480 & .1442 & 98.75 $\pm$ 0.04\\
 $WCT^2$ & 10.11 
 & .3635 & \textbf{.0001} & \textbf{.0019} & 88.23 $\pm$ 0.21 & \textbf{.0074} & \textbf{.0263} & 66.07 $\pm$ 0.39 & \textbf{.0013} & \underline{.0270} & 98.22 $\pm$ 0.06 & \textbf{.0008} & \underline{.0147} & 100 $\pm$ 0 \\
 Splice & 1.04 
 & 762 & .0167 & .0174 & 93.75 $\pm$ 0.11 & .0767 & \textbf{.0263} & 69.8 $\pm$ 0.38 & .0521 & .0392 & 72.23 $\pm$ 0.37 & .4699 & .5365 & 78.27 $\pm$ 0.34 \\
SpliceNet
& 54.43 
& \textbf{.0892} & \underline{.0035} & \underline{.0078} & - & \underline{.0135} & \underline{.0379} & - & \underline{.0037} & \textbf{.0144} & - & \underline{.0039} & \textbf{.0107} & - \\
\hline
\end{tabular}
\caption{For each baseline, we report: model size and runtime (measures for $512px$ images on RTX6000 GPU). For each dataset, we report reconstruction error measured by LPIPS$\downarrow$, MSE$\downarrow$, and human perceptual evaluation results, measured by the percentage of judgments in our favor (mean, std).}\vspace{-0.55cm}
\label{table:reconstruction}
\end{table*}}

{\small \begin{table}
\centering
    \renewcommand{\tabcolsep}{4pt}
\begin{tabular}{c | c | c | c | c | c} 
 \hline
 & Kim et al. & SA &  WCT$^2$ & Splice (Ours)& SpliceNet (Ours) \\ 
 \hline
 \emph{AFHQ} & 0.826 & 0.773 & 0.516 & 0.677 & \textbf{0.225} \\ 
 \hline
 \emph{Oxford-102}  & 0.64 & 0.933 & 0.819 & 0.518 & \textbf{0.507} \\ 
 \hline
 \emph{SD-Dogs}  & 0.849 & 0.612 & 0.568 & 0.462 & \textbf{0.435} \\ 
 \hline
 \emph{SD-Horses}  & 0.809 & \textbf{0.568} & 0.775 & 0.869 & 0.577 \\ 
 \hline
\end{tabular}
\caption{\label{table:sifid} We report the average SI-FID computed over 100 random pairs from each dataset. Lower is better.}\vspace{-0.65cm}
\end{table}}


For evaluating \textbf{SpliceNet}, we perform the survey using  80 image-pairs from all datasets. We collected 6500 user judgments w.r.t. existing baselines. Table \ref{table:reconstruction} reports the percentage of votes in our favor.  As seen, our method outperforms all baselines across all datasets, especially in the Internet datasets (\emph{SD-Dogs}, \emph{SD-Horses}), which highlights our performance in challenging settings. 

{\small \begin{table} 
\centering
    \renewcommand{\tabcolsep}{4pt}
\begin{tabular}{c | c | c | c | c | c} 
 \hline
 & Kim et al. & SA &  WCT$^2$ & Splice (Ours) & SpliceNet (Ours) \\ 
\hline
\emph{AFHQ}  & 0.953  & \textbf{0.967}  & 0.961  & 0.883  & \textbf{0.967} \\
\hline
\emph{SD-Dogs} & 0.938  & 0.801  & \textbf{0.954}  & 0.863  &\textbf{0.954} \\
\hline
\emph{SD-Horses} & 0.928  & 0.861  & \textbf{0.948}  & 0.867  & 0.940 \\
\hline
\end{tabular}
\caption{\label{table:splicenet-segmentation}We extract semantic segmentation maps of objects of interest in the content and stylized images. Mean IoU  over 100 images are reported for each dataset.}\vspace{-0.55cm}
\end{table}}

\begin{table} 
\centering
    \begin{adjustbox}{max width=\linewidth}
    \renewcommand{\tabcolsep}{4pt}
\begin{tabular}{c | c | c | c} 
 \hline
 & SA & STROTSS & WCT$^2$ \\ 
 \hline
 Wild-Pairs & - & \textbf{79.0} $\pm$ 13.0 & \textbf{83.1} $\pm$ 14.9 \\ 
 \hline
 mountains & \textbf{56.3} $\pm$ 10.0 & \textbf{58.8} $\pm$ 14.2 & \textbf{60.3} $\pm$ 12.1 \\
 \hline
 AFHQ & \textbf{71.8} $\pm$ 7.7 & \textbf{59.7} $\pm$ 15.3 & \textbf{61.0} $\pm$ 18.3 \\
 \hline
\end{tabular}
\end{adjustbox}
\caption{\textbf{Splice AMT perceptual evaluation}. We report results on AMT surveys evaluating the task of appearance transfer across semantically related scenes/objects (see Sec.~\ref{sec:quantitive}). For each dataset and a baseline, we report the percentage of judgments in our favor (mean, std). Our method outperforms   all  baselines: GAN-based, SA~\cite{park2020swapping}, and style transfer methods, STROTSS~\cite{kolkin2019style}, and WCT$^2$~\cite{Yoo_2019_ICCV}.}

\label{table:amt}
\end{table}

\begin{table}[]
    \centering
    \renewcommand{\tabcolsep}{4pt}
    \begin{adjustbox}{max width=\linewidth}
    \begin{tabular}{c|c|c|c|c} 
    \hline
        & SA & STROTSS & WCT$^2$ & Splice (Ours) \\ 
        \hline
        Wild-Pairs & - & 0.83±0.11 & \textbf{0.89±0.06} & 0.88±0.06\\ 
        \hline
        mountains & 0.91±0.07 & 0.94±0.12 & \textbf{0.96±0.82} & 0.95±0.10\\
        \hline
    \end{tabular}
    \end{adjustbox}
    \caption{\textbf{Mean IoU of output images with respect to the input structure images}. We extract semantic segmentation maps using Mask-RCNN \cite{he2017mask} for the Wild-Pairs collection, and \cite{zhou2018semantic_seg} for the mountains collection.} 
    \label{tab:seg_results} \afterfigure
\end{table}

\begin{figure} 
    \centering
    \includegraphics[width=.48\textwidth]{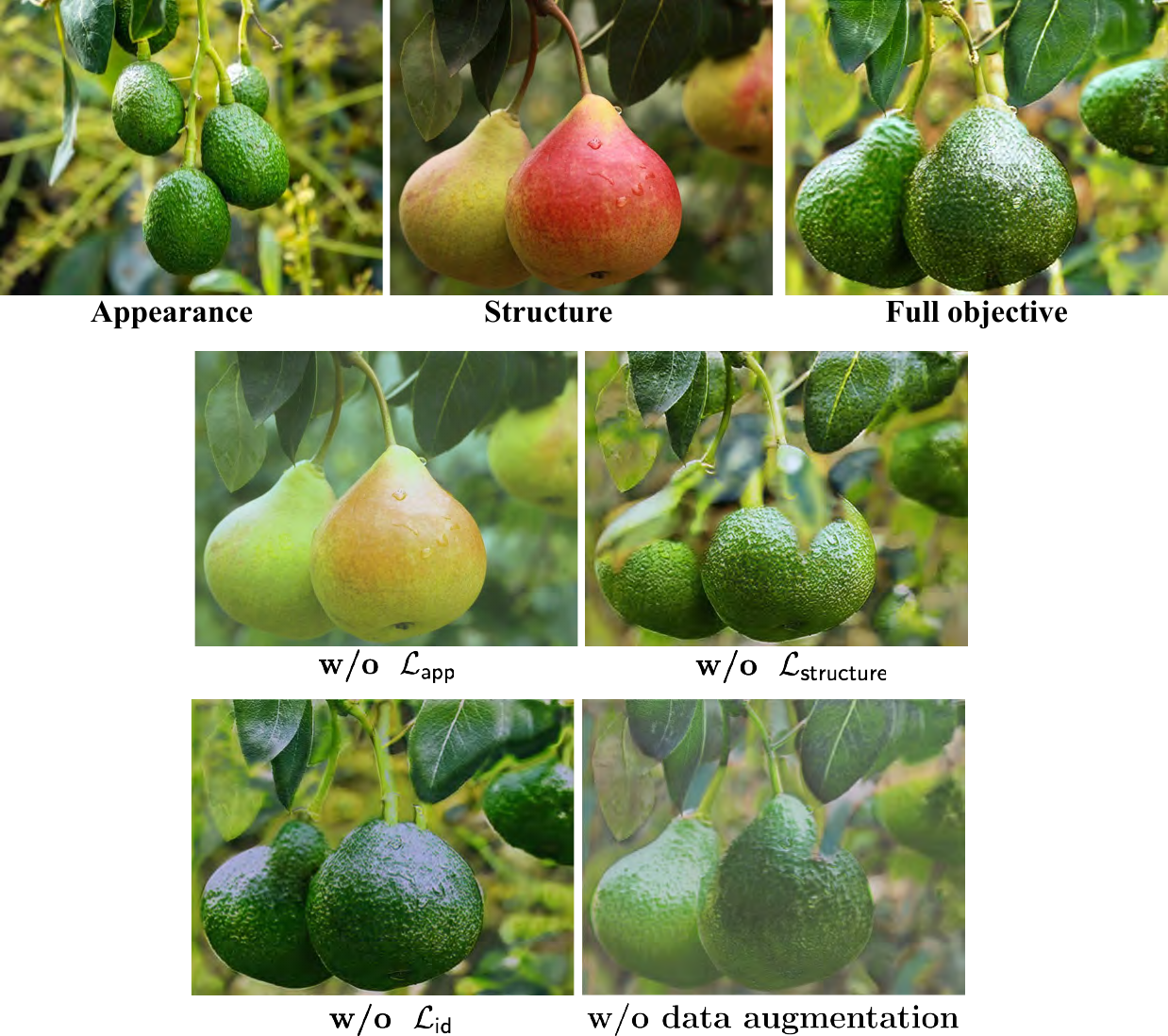} \vspace{-0.7cm}
    \caption{{\bf Loss and data augmentation ablations.} Splice ablation results of specific loss terms and the data augmentation. When one of our loss terms is removed, the model fails to map the target appearance, preserve the input structure, or maintain fine details. Without dataset augmentation, while the semantic association is largely maintained, the visual quality is significantly decreased. See Sec.~\ref{sec:ablation} for more details.}\afterfigure
    \label{fig:ablation}
\end{figure}

\begin{figure}
    \centering
    \vspace{0.5cm}
    \includegraphics[width=.5\textwidth]{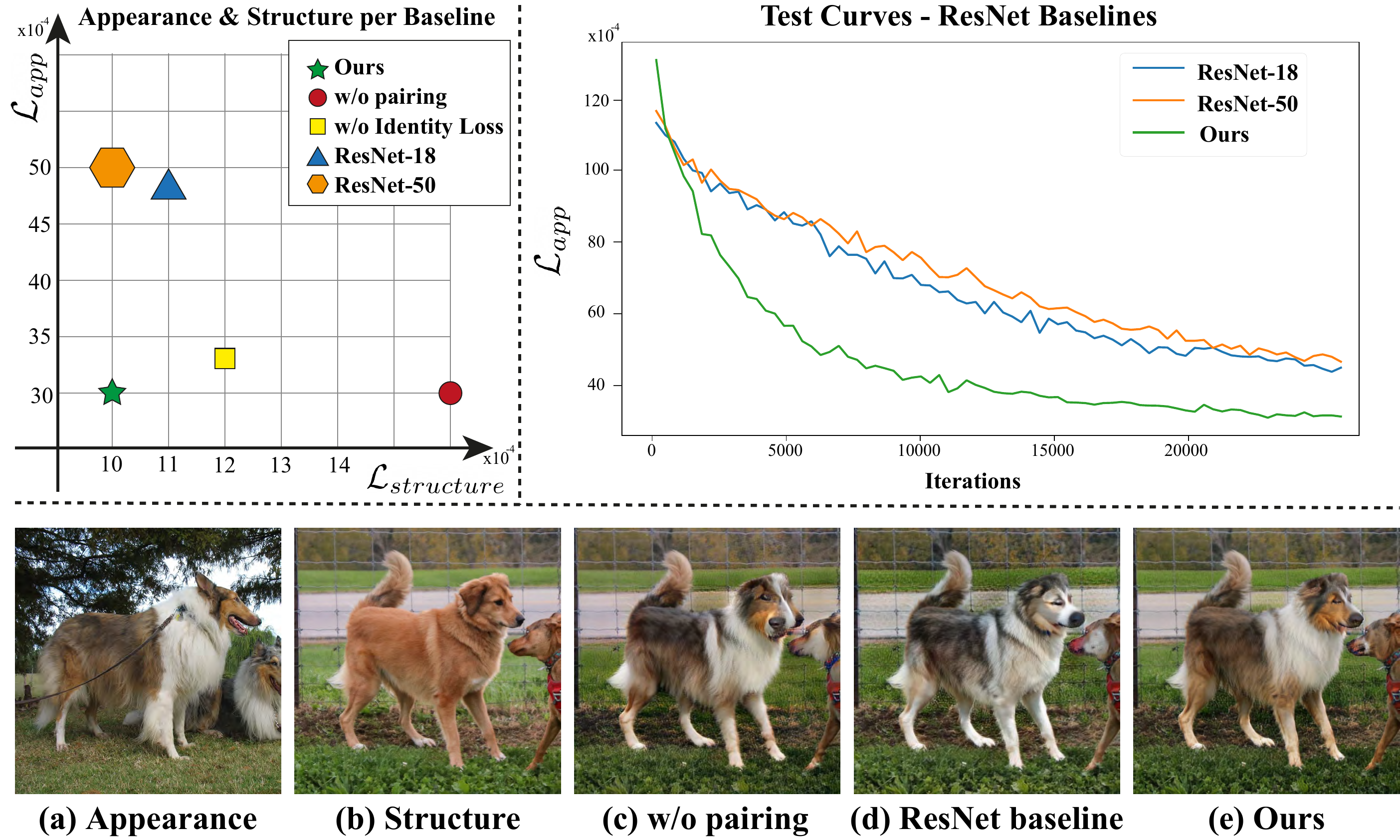} \vspace{-0.7cm}
    \caption{{\bf SpliceNet ablations.} Top left: we plot style/content losses for several baselines, including our model trained w/o data distillation, and w/o  conditioning on the style token (ResNet baseline). Top right: test loss curves computed during training for our framework vs. the ResNet baselines. Bottom: qualitative comparison of a representative pair. See Sec.~\ref{sec:ablation} for details.} 
    \label{fig:splicenet-ablation}  \afterfigure
\end{figure}

\begin{figure*}[t!]
    \centering
    \includegraphics[width=\textwidth]{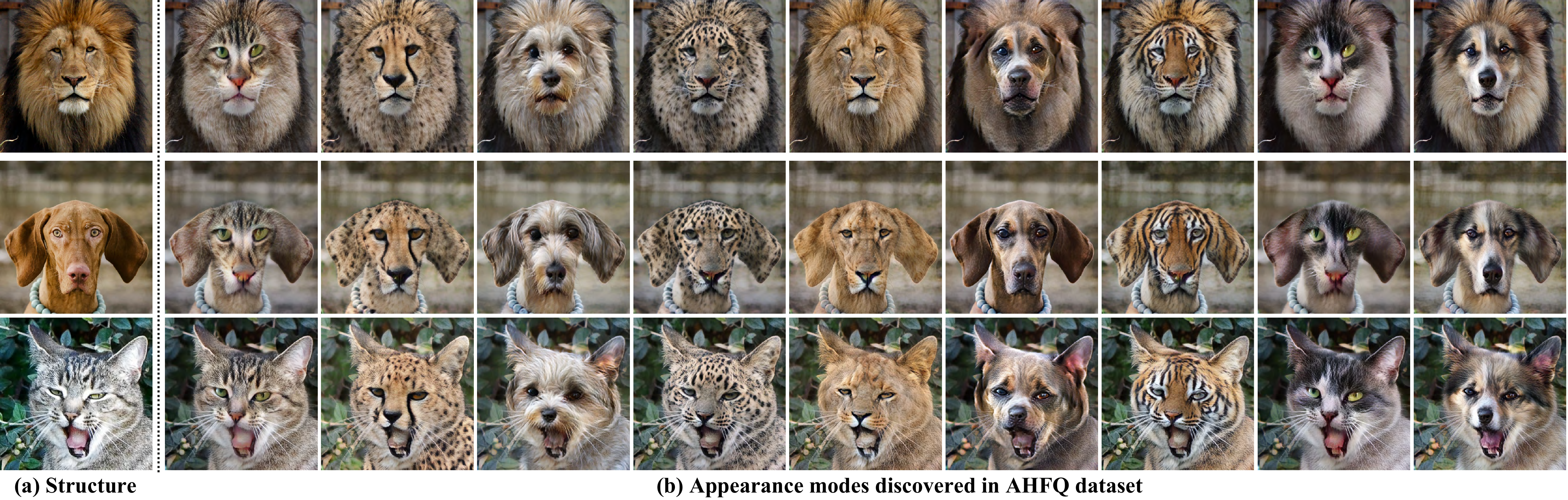}\vspace{-0.3cm}
    \caption{{\bf Appearance modes} are discovered by clustering the \cls \ token across all AFHQ training set. (b) We transfer each of the discovered appearance modes to test structure images (a). }\afterfigure
    \label{fig:style_modes}
\end{figure*}

\begin{figure}
    \centering
    \includegraphics[width=.98\linewidth]{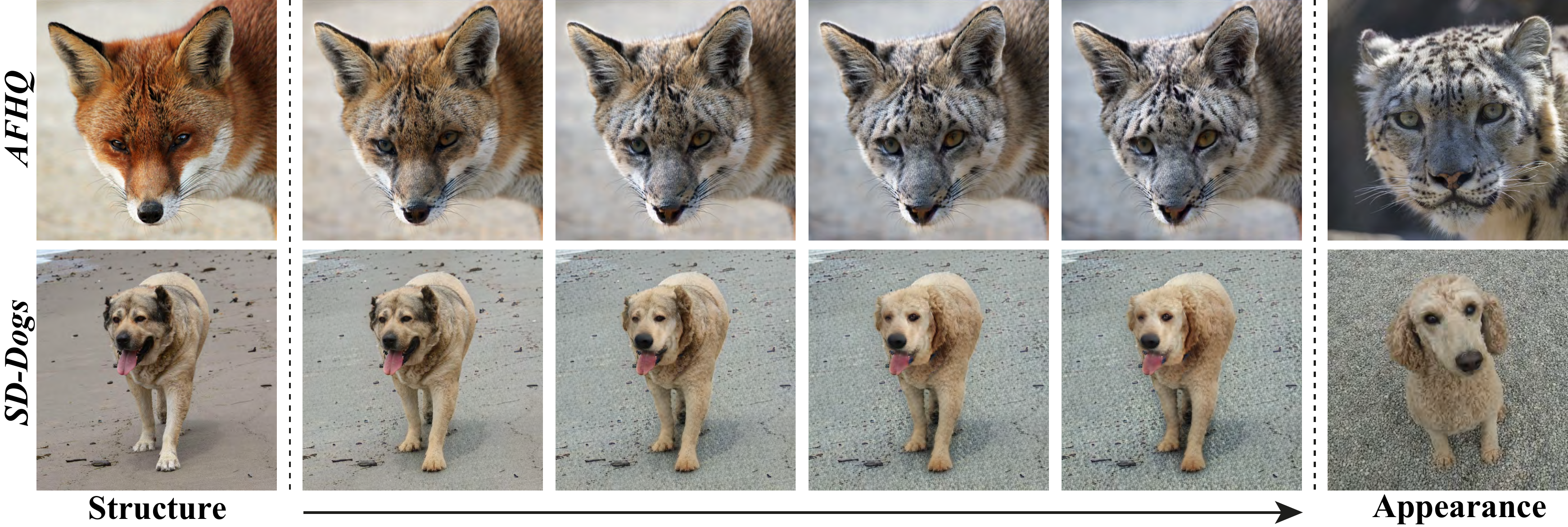} \vspace{-0.4cm}
    \caption{{\bf Appearance interpolation} Controlling stylization extent via interpolation in \cls \ token space.}\afterfigure
    \label{fig:interpolation}
\end{figure}


\myparagraph{Semantic layout preservation.} A key property of our method is the ability to preserve the semantic layout of the scene (while significantly changing the appearance of objects). We demonstrate this through the following evaluation.  We run semantic segmentation off-the-shelf model (e.g., MaskRCNN~\cite{he2017mask}) to compute object masks for the input structure images and  our results.

Table \ref{tab:seg_results} reports IoU for Splice and the baselines. Splice better preserves the scene layout than SA and STROTSS, and is the closet competitor to WCT$^2$ which only modifies colors, and as expected, achieves the highest IoU.

We perform the same evaluation protocol on SpliceNet and its competitors. We consider the objects relevant to our datasets for which clean and robust segmentation masks could be obtained (cats, dogs, horses).  Table~\ref{table:splicenet-segmentation} reports  the average intersection over union (IoU) between the masks computed for the content images and the corresponding stylized results.
SpliceNet achieves higher (better) IoU  than \citeauthor{styleDiscriminator} and Splice and is the closet competitor to WCT$^2$, which achieves the highest IoU as it only modifies colors.

\myparagraph{Reconstruction.}
When the input appearance and structure images are identical, we expect any appearance transfer method to  \emph{reconstruct} the input image.
Table~\ref{table:reconstruction} reports mean squared error (MSE), and LPIPS~\cite{lpips} computed between the input and the reconstructed image. 
Naturally, WCT$^2$ excels in most datasets since it does not synthesize new textures or modify shapes. 
SpliceNet surpasses all other methods, including state-of-the-art GAN-based methods, by an order of magnitude.

\subsection{Ablation}
\label{sec:ablation}

We ablate the loss terms and design choices in our proposed frameworks.

\paragraph{Loss terms.} We ablate the different loss terms in our objective function by qualitatively comparing the results when trained with the full objective (Eq.~\ref{eq:4}), and with a specific loss removed. The results are shown in Fig.~\ref{fig:ablation}. As can be seen, without the {\bf appearance loss} (w/o $\mathcal{L}_{\mathsf{app}}$), Splice fails to map the target appearance, but only slightly modifies the colors of the input structure image due to the identity loss. That is, the identity loss encourages the model to learn an identity when it is fed with the target appearance image, and therefore even without the appearance loss some appearance supervision is available.
Without the {\bf structure loss}~(w/o $\mathcal{L}_{\mathsf{structure}}$), the model outputs an image with the desired appearance, but fails to fully preserve the structure of the input image, as can be seen by the distorted shape of the pears. Lastly, we observe that the {\bf identity loss} encourages the model to pay more attention to fine details both in terms of appearance and structure, e.g., the fine texture details of the avocado are refined.

\paragraph{Dataset augmentation.} We ablate the usage of dataset augmentation in Splice. In this case, the network solves a test-time optimization problem between two images rather than learning to map between many internal examples. As can be seen in Fig.~\ref{fig:ablation}, without data augmentation, the semantic association is largely preserved, however, the realism and visual quality of Splice are significantly decreased.

~

For SpliceNet, we ablate our key design choices by considering these baselines:
\paragraph{Input \cls \ token vs. input appearance image} To demonstrate the effectiveness of directly using DINO-ViT's \cls \ token as input, we consider a baseline architecture that takes as input $I_t$, the appearance \emph{image}. Specifically, we use an off-the-shelf ResNet backbone \cite{he2016deep} to map $I_t$ into a global appearance vector which is mapped to modulation parameters via learnable affine transformations.

\paragraph{No structure/appearance pair distillation.}  We show the importance of our data curation (Sec.~\ref{sec:data}) by training a model on random image pairs. 

Figure~\ref{fig:splicenet-ablation}(bottom) shows a qualitative comparison to the above baselines on a sample pair (see SM for more examples). As seen in  Fig.~\ref{fig:splicenet-ablation}(d), without conditioning the model on the \cls \ token, the results suffer from visual artifacts and the model could not deviate much from the original texture. As seen in Fig.~\ref{fig:splicenet-ablation}(c), a model trained without pairs distillation (w/o pairing) can still synthesize textures matching the target appearance, yet fail to preserve the semantic content. 

We quantify these results as follows: We randomly sample input pairs from \emph{SD-Dogs} test set, and compute the average structure and appearance losses (Eq.~\ref{eq:4}). Figure~\ref{fig:splicenet-ablation}(top left) reports the results for all baselines, and validate the expected trends. 

Figure ~\ref{fig:splicenet-ablation} (top right) shows the learning curves on \emph{SD-Dogs} test set for the different CNN backbones. As can be seen, directly conditioning on the \cls \ token results in faster convergence, and lower appearance loss. Fig.~\ref{fig:splicenet-ablation}(d).

\subsection{Manipulation in \cls{} Token Space}

Directly conditioning SpliceNet on the \cls \ token space not only benefits the appearance transfer and training convergence, but also enables applications of appearance transfer by performing manipulations in the \cls \ token space. Specifically, we perform interpolation between the structure and appearance \cls \ tokens to control the stylization extent, and detect appearance modes by performing K-means on the \cls \ tokens of the dataset.

\paragraph{Appearance Interpolation.} We can control the extent of stylization by feeding to our model interpolating the style tokens of the style and content images, i.e., $t_i = \alpha_i t_{\mathcls}^{L}(I_t) + (1-\alpha_i)t_{\mathcls}^{L}(I_s)$. Sample examples are shown in Fig.~\ref{fig:interpolation} and more included in SM. 

\paragraph{Detecting and Visualizing Appearance Modes.} We automatically discover representative appearances, i.e., \emph{appearance modes} in the data. To do so, we extract the \cls \ token for all images in the training set, and apply K-means, where the centroids are used as our \emph{appearance modes}. 
We visualize the modes by using each as the input \cls \ token to SpliceNet, along with a structure image . Figure~\ref{fig:style_modes} shows nine such modes automatically discovered for AFHQ training set, transferred to test set structure images.  More examples of appearance modes are in SM.

\section{Limitations}

\label{sec:limitations}

The performance of our frameworks depends on the internal representation learned by DINO-ViT, and is therefore limited in several aspects.

\begin{figure} 
    \centering
    \includegraphics[width=.45\textwidth]{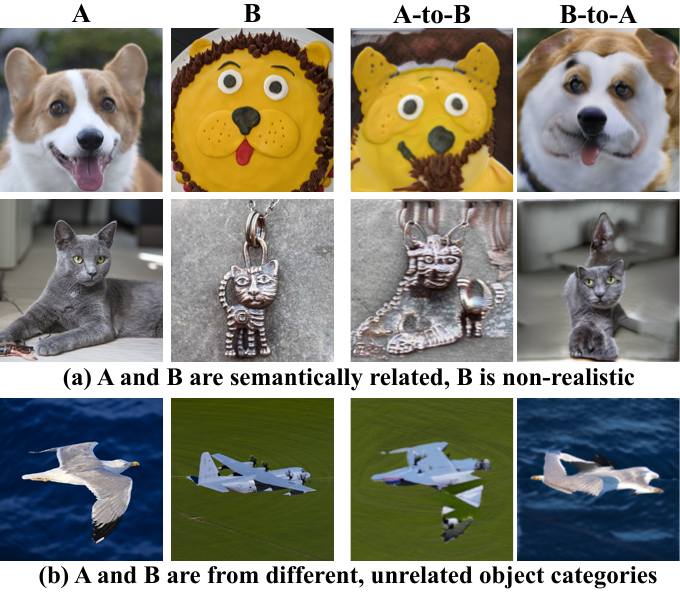}
    \caption{{\bf Splice limitations.} (a) Objects in the input images (A-B) are semantically related, yet B is non-realistic. (b) Objects are from unrelated object categories. See Sec.~\ref{sec:limitations} for discussion.}\afterfigure
    \label{fig:limitation}
\end{figure}

\begin{figure}
    \centering
    \includegraphics[width=.45\textwidth]{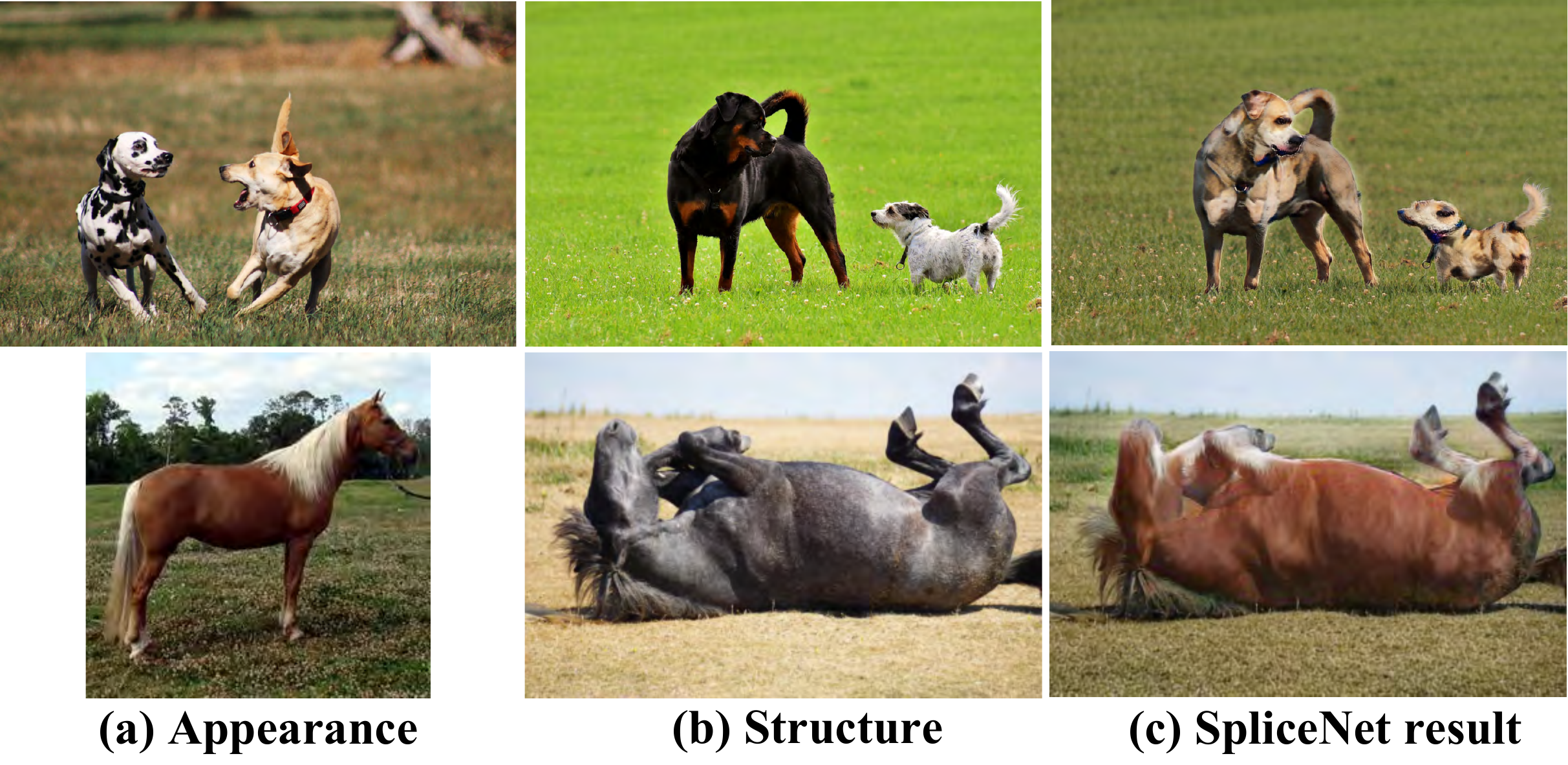} \vspace{-0.4cm}
    \caption{{\bf SpliceNet limitations.} Top: the target style contains multiple foreground objects, where our result depict a single blended style. Bottom: under extreme pose variations, our method may fail to establish accurate semantic association.} \afterfigure
    \label{fig:splicenet-limitations}
\end{figure}

First, our frameworks are limited by the features' expressiveness. For example, our method can fail to make the correct semantic association in case the DINO-ViT representation fails to capture it. Figure~\ref{fig:limitation} shows a few such cases for Splice: (a) objects are semantically related but one image is highly non-realistic (and thus out of distribution for \dinovit). For some regions, Splice successfully transfers the appearance but for some others it fails. In the cat example, we can see that in B-to-A result, the face and the body of the cat are  nicely mapped, yet Splice fails to find a semantic correspondence for the rings, and we get a wrong mapping of the ear from image A. In (b), Splice does not manage to semantically relate a bird to an airplane. We also found that the \cls \ token cannot faithfully capture distinct appearances of \emph{multiple foreground} objects, but rather captures a joint blended appearance. This can be seen in Fig.~\ref{fig:splicenet-limitations}(top), where SpliceNet transfers the ``averaged" appearance to the structure image.

Second, if the structure and appearance test pair contains extreme pose variation, our method may fail to establish correct semantic association, as seen in Fig.~\ref{fig:splicenet-limitations}(bottom).

Third, Splice is restricted to observing only a single image pair and is subject to optimization instabilities, which can lead to incorrect semantic association or poor visual quality, as discussed in Sec. \ref{sec:comparison}. SpliceNet overcomes these limitations due to being trained on a dataset, which makes it more robust to challenging inputs and enhances the visual quality.

Finally, DINO-ViT has been trained on ImageNet and thus our models can be trained on domains that are well-represented in DINO-ViT's training data. This can be tackled by re-training or fine-tuning DINO-ViT on other domains.

\section{Conclusions}
\label{sec:conclusions}

We tackled a new problem setting in the context of style/appearance transfer: semantically transferring appearance across related objects in two in-the-wild natural images, without any user guidance. Our approach demonstrates the power of DINO-ViT as an external semantic prior, and the effectiveness of utilizing it to establish our training losses -- we show how structure and appearance information can be disentangled from an input image, and then spliced together in a semantically meaningful way in the space of ViT features, through a generation process.
We propose two frameworks of semantic appearance transfer based on our perceptual losses: (i) Splice, which is a generator trained on a single and arbitrary structure-appearance input pair, and (ii) SpliceNet, a feed-forward generator trained on a domain-specific dataset. Direct conditioning on ViT features boosts the performance of SpliceNet in terms of visual quality and convergence rate. We further showed how to distill suitable training data for SpliceNet from noisy diverse image collections. 

We demonstrated that our method can be applied on a variety of challenging input pairs across domains, in diverse poses and multiplicity of objects, and can produce high-quality result without any adversarial training. Through extensive evaluation, we showed that our frameworks, trained with simple perceptual losses, excel state-of-the-art GAN-based methods.

Our evaluations demonstrate that SpliceNet surpasses Splice in terms of visual quality, and is orders of magnitude faster, enabling real-time semantic appearance transfer. Moreover, Splice is limited to observing only a single test-time pair and is subject to instabilities during its optimization process, which may lead to incorrect semantic association and poor visual quality. On the other hand, since SpliceNet is trained on a dataset of semantically related image pairs, it results in a better semantic association and generalization, and is more robust to challenging input pairs. However, SpliceNet is trained on a domain-specific dataset, hence is limited to input images from that domain. In contrast, Splice works on arbitrary, in-the-wild input pairs, without being restricted to a particular domain.

We believe that our work unveils the potential of self-supervised representation learning not only for discriminative tasks such as image classification, but also for learning more powerful generative models.

~\newline

\subsubsection*{Acknowledgments:} We would like to thank Meirav Galun for her insightful comments and discussion.
This project received funding from the Israeli Science Foundation (grant 2303/20), and the Carolito Stiftung. Dr Bagon is a Robin Chemers Neustein Artificial Intelligence Fellow.

~\newline




\bibliographystyle{ACM-Reference-Format}
\bibliography{sample-base}


\begin{thebibliography}{58}


\ifx \showCODEN    \undefined \def \showCODEN     #1{\unskip}     \fi
\ifx \showDOI      \undefined \def \showDOI       #1{#1}\fi
\ifx \showISBNx    \undefined \def \showISBNx     #1{\unskip}     \fi
\ifx \showISBNxiii \undefined \def \showISBNxiii  #1{\unskip}     \fi
\ifx \showISSN     \undefined \def \showISSN      #1{\unskip}     \fi
\ifx \showLCCN     \undefined \def \showLCCN      #1{\unskip}     \fi
\ifx \shownote     \undefined \def \shownote      #1{#1}          \fi
\ifx \showarticletitle \undefined \def \showarticletitle #1{#1}   \fi
\ifx \showURL      \undefined \def \showURL       {\relax}        \fi
\providecommand\bibfield[2]{#2}
\providecommand\bibinfo[2]{#2}
\providecommand\natexlab[1]{#1}
\providecommand\showeprint[2][]{arXiv:#2}

\bibitem[Amir et~al\mbox{.}(2022)]%
        {amir2021deep}
\bibfield{author}{\bibinfo{person}{Shir Amir}, \bibinfo{person}{Yossi Gandelsman}, \bibinfo{person}{Shai Bagon}, {and} \bibinfo{person}{Tali Dekel}.} \bibinfo{year}{2022}\natexlab{}.
\newblock \showarticletitle{Deep ViT Features as Dense Visual Descriptors}.
\newblock \bibinfo{journal}{\emph{ECCVW What is Motion For?}} (\bibinfo{year}{2022}).
\newblock


\bibitem[Beech(2005)]%
        {splicing}
\bibfield{author}{\bibinfo{person}{Frank Beech}.} \bibinfo{year}{2005}\natexlab{}.
\newblock \showarticletitle{Splicing Ropes Illustrated}.
\newblock \bibinfo{journal}{\emph{CCCBR}} (\bibinfo{year}{2005}).
\newblock


\bibitem[Benaim et~al\mbox{.}(2021)]%
        {DBLP:journals/cgf/BenaimMBW21}
\bibfield{author}{\bibinfo{person}{Saguy Benaim}, \bibinfo{person}{Ron Mokady}, \bibinfo{person}{Amit Bermano}, {and} \bibinfo{person}{Lior Wolf}.} \bibinfo{year}{2021}\natexlab{}.
\newblock \showarticletitle{Structural Analogy from a Single Image Pair}.
\newblock \bibinfo{journal}{\emph{Comput. Graph. Forum}} (\bibinfo{year}{2021}).
\newblock


\bibitem[Caron et~al\mbox{.}(2021)]%
        {dino}
\bibfield{author}{\bibinfo{person}{Mathilde Caron}, \bibinfo{person}{Hugo Touvron}, \bibinfo{person}{Ishan Misra}, \bibinfo{person}{Herv{\'e} J{\'e}gou}, \bibinfo{person}{Julien Mairal}, \bibinfo{person}{Piotr Bojanowski}, {and} \bibinfo{person}{Armand Joulin}.} \bibinfo{year}{2021}\natexlab{}.
\newblock \showarticletitle{Emerging properties in self-supervised vision transformers}. In \bibinfo{booktitle}{\emph{Proceedings of the IEEE/CVF International Conference on Computer Vision}}.
\newblock


\bibitem[Champandard(2016)]%
        {Champandard16}
\bibfield{author}{\bibinfo{person}{Alex~J. Champandard}.} \bibinfo{year}{2016}\natexlab{}.
\newblock \showarticletitle{Semantic Style Transfer and Turning Two-Bit Doodles into Fine Artworks}.
\newblock \bibinfo{journal}{\emph{arXiv}} (\bibinfo{year}{2016}).
\newblock


\bibitem[Chen and Schmidt(2016)]%
        {chen2016fast}
\bibfield{author}{\bibinfo{person}{Tian~Qi Chen} {and} \bibinfo{person}{Mark Schmidt}.} \bibinfo{year}{2016}\natexlab{}.
\newblock \showarticletitle{Fast patch-based style transfer of arbitrary style}.
\newblock \bibinfo{journal}{\emph{arXiv preprint arXiv:1612.04337}} (\bibinfo{year}{2016}).
\newblock


\bibitem[Choi et~al\mbox{.}(2020)]%
        {Choi_2020_CVPR}
\bibfield{author}{\bibinfo{person}{Yunjey Choi}, \bibinfo{person}{Youngjung Uh}, \bibinfo{person}{Jaejun Yoo}, {and} \bibinfo{person}{Jung-Woo Ha}.} \bibinfo{year}{2020}\natexlab{}.
\newblock \showarticletitle{StarGAN v2: Diverse Image Synthesis for Multiple Domains}. In \bibinfo{booktitle}{\emph{Proceedings of the IEEE/CVF Conference on Computer Vision and Pattern Recognition (CVPR)}}.
\newblock


\bibitem[Cohen and Wolf(2019)]%
        {cohen2019bidirectional}
\bibfield{author}{\bibinfo{person}{Tomer Cohen} {and} \bibinfo{person}{Lior Wolf}.} \bibinfo{year}{2019}\natexlab{}.
\newblock \showarticletitle{Bidirectional one-shot unsupervised domain mapping}. In \bibinfo{booktitle}{\emph{Proceedings of the IEEE/CVF International Conference on Computer Vision}}.
\newblock


\bibitem[Dekel et~al\mbox{.}(2015)]%
        {dekel2015best}
\bibfield{author}{\bibinfo{person}{Tali Dekel}, \bibinfo{person}{Shaul Oron}, \bibinfo{person}{Michael Rubinstein}, \bibinfo{person}{Shai Avidan}, {and} \bibinfo{person}{William~T Freeman}.} \bibinfo{year}{2015}\natexlab{}.
\newblock \showarticletitle{Best-buddies similarity for robust template matching}. In \bibinfo{booktitle}{\emph{Proceedings of the IEEE conference on computer vision and pattern recognition}}. \bibinfo{pages}{2021--2029}.
\newblock


\bibitem[Dosovitskiy et~al\mbox{.}(2021)]%
        {vit}
\bibfield{author}{\bibinfo{person}{Alexey Dosovitskiy}, \bibinfo{person}{Lucas Beyer}, \bibinfo{person}{Alexander Kolesnikov}, \bibinfo{person}{Dirk Weissenborn}, \bibinfo{person}{Xiaohua Zhai}, \bibinfo{person}{Thomas Unterthiner}, \bibinfo{person}{Mostafa Dehghani}, \bibinfo{person}{Matthias Minderer}, \bibinfo{person}{Georg Heigold}, \bibinfo{person}{Sylvain Gelly}, \bibinfo{person}{Jakob Uszkoreit}, {and} \bibinfo{person}{Neil Houlsby}.} \bibinfo{year}{2021}\natexlab{}.
\newblock \showarticletitle{An Image is Worth 16x16 Words: Transformers for Image Recognition at Scale}.
\newblock \bibinfo{journal}{\emph{ICLR}} (\bibinfo{year}{2021}).
\newblock


\bibitem[Dumoulin et~al\mbox{.}(2017)]%
        {dumoulin2016learned}
\bibfield{author}{\bibinfo{person}{Vincent Dumoulin}, \bibinfo{person}{Jonathon Shlens}, {and} \bibinfo{person}{Manjunath Kudlur}.} \bibinfo{year}{2017}\natexlab{}.
\newblock \showarticletitle{A Learned Representation For Artistic Style}. In \bibinfo{booktitle}{\emph{International Conference on Learning Representations}}.
\newblock


\bibitem[Gatys et~al\mbox{.}(2017)]%
        {GatysEBHS17}
\bibfield{author}{\bibinfo{person}{Leon~A. Gatys}, \bibinfo{person}{Alexander~S. Ecker}, \bibinfo{person}{Matthias Bethge}, \bibinfo{person}{Aaron Hertzmann}, {and} \bibinfo{person}{Eli Shechtman}.} \bibinfo{year}{2017}\natexlab{}.
\newblock \showarticletitle{Controlling Perceptual Factors in Neural Style Transfer}.
\newblock


\bibitem[He et~al\mbox{.}(2017)]%
        {he2017mask}
\bibfield{author}{\bibinfo{person}{Kaiming He}, \bibinfo{person}{Georgia Gkioxari}, \bibinfo{person}{Piotr Doll{\'a}r}, {and} \bibinfo{person}{Ross Girshick}.} \bibinfo{year}{2017}\natexlab{}.
\newblock \showarticletitle{Mask r-cnn}. In \bibinfo{booktitle}{\emph{Proceedings of the IEEE international conference on computer vision}}.
\newblock


\bibitem[He et~al\mbox{.}(2016)]%
        {he2016deep}
\bibfield{author}{\bibinfo{person}{Kaiming He}, \bibinfo{person}{Xiangyu Zhang}, \bibinfo{person}{Shaoqing Ren}, {and} \bibinfo{person}{Jian Sun}.} \bibinfo{year}{2016}\natexlab{}.
\newblock \showarticletitle{Deep residual learning for image recognition}. In \bibinfo{booktitle}{\emph{Proceedings of the IEEE conference on computer vision and pattern recognition}}. \bibinfo{pages}{770--778}.
\newblock


\bibitem[Huang and Belongie(2017)]%
        {adain}
\bibfield{author}{\bibinfo{person}{Xun Huang} {and} \bibinfo{person}{Serge Belongie}.} \bibinfo{year}{2017}\natexlab{}.
\newblock \showarticletitle{Arbitrary style transfer in real-time with adaptive instance normalization}. In \bibinfo{booktitle}{\emph{Proceedings of the IEEE international conference on computer vision}}. \bibinfo{pages}{1501--1510}.
\newblock


\bibitem[Isola et~al\mbox{.}(2017)]%
        {8100115}
\bibfield{author}{\bibinfo{person}{Phillip Isola}, \bibinfo{person}{Jun-Yan Zhu}, \bibinfo{person}{Tinghui Zhou}, {and} \bibinfo{person}{Alexei~A. Efros}.} \bibinfo{year}{2017}\natexlab{}.
\newblock \showarticletitle{Image-to-Image Translation with Conditional Adversarial Networks}. In \bibinfo{booktitle}{\emph{2017 IEEE Conference on Computer Vision and Pattern Recognition (CVPR)}}.
\newblock


\bibitem[Jing et~al\mbox{.}(2020)]%
        {JingYFYYS20}
\bibfield{author}{\bibinfo{person}{Yongcheng Jing}, \bibinfo{person}{Yezhou Yang}, \bibinfo{person}{Zunlei Feng}, \bibinfo{person}{Jingwen Ye}, \bibinfo{person}{Yizhou Yu}, {and} \bibinfo{person}{Mingli Song}.} \bibinfo{year}{2020}\natexlab{}.
\newblock \showarticletitle{Neural Style Transfer: {A} Review}.
\newblock \bibinfo{journal}{\emph{{IEEE} Trans. Vis. Comput. Graph.}} (\bibinfo{year}{2020}).
\newblock


\bibitem[Johnson et~al\mbox{.}(2016)]%
        {johnson2016perceptual}
\bibfield{author}{\bibinfo{person}{Justin Johnson}, \bibinfo{person}{Alexandre Alahi}, {and} \bibinfo{person}{Li Fei-Fei}.} \bibinfo{year}{2016}\natexlab{}.
\newblock \showarticletitle{Perceptual losses for real-time style transfer and super-resolution}. In \bibinfo{booktitle}{\emph{European conference on computer vision}}. Springer, \bibinfo{pages}{694--711}.
\newblock


\bibitem[Karras et~al\mbox{.}(2020)]%
        {stylegan2}
\bibfield{author}{\bibinfo{person}{Tero Karras}, \bibinfo{person}{Samuli Laine}, \bibinfo{person}{Miika Aittala}, \bibinfo{person}{Janne Hellsten}, \bibinfo{person}{Jaakko Lehtinen}, {and} \bibinfo{person}{Timo Aila}.} \bibinfo{year}{2020}\natexlab{}.
\newblock \showarticletitle{Analyzing and Improving the Image Quality of {StyleGAN}}. In \bibinfo{booktitle}{\emph{Proc. CVPR}}.
\newblock


\bibitem[Kim et~al\mbox{.}(2022)]%
        {styleDiscriminator}
\bibfield{author}{\bibinfo{person}{Kunhee Kim}, \bibinfo{person}{Sanghun Park}, \bibinfo{person}{Eunyeong Jeon}, \bibinfo{person}{Taehun Kim}, {and} \bibinfo{person}{Daijin Kim}.} \bibinfo{year}{2022}\natexlab{}.
\newblock \showarticletitle{A Style-aware Discriminator for Controllable Image Translation}. In \bibinfo{booktitle}{\emph{IEEE/CVF Conference on Computer Vision and Pattern Recognition}}.
\newblock


\bibitem[Kim et~al\mbox{.}(2020)]%
        {kim2020deformable}
\bibfield{author}{\bibinfo{person}{Sunnie~SY Kim}, \bibinfo{person}{Nicholas Kolkin}, \bibinfo{person}{Jason Salavon}, {and} \bibinfo{person}{Gregory Shakhnarovich}.} \bibinfo{year}{2020}\natexlab{}.
\newblock \showarticletitle{Deformable style transfer}.
\newblock


\bibitem[Kim et~al\mbox{.}(2017)]%
        {kim2017learning}
\bibfield{author}{\bibinfo{person}{Taeksoo Kim}, \bibinfo{person}{Moonsu Cha}, \bibinfo{person}{Hyunsoo Kim}, \bibinfo{person}{Jung~Kwon Lee}, {and} \bibinfo{person}{Jiwon Kim}.} \bibinfo{year}{2017}\natexlab{}.
\newblock \showarticletitle{Learning to discover cross-domain relations with generative adversarial networks}. In \bibinfo{booktitle}{\emph{International Conference on Machine Learning}}. PMLR.
\newblock


\bibitem[Kingma and Ba(2015)]%
        {DBLP:journals/corr/KingmaB14}
\bibfield{author}{\bibinfo{person}{Diederik~P. Kingma} {and} \bibinfo{person}{Jimmy Ba}.} \bibinfo{year}{2015}\natexlab{}.
\newblock \showarticletitle{Adam: {A} Method for Stochastic Optimization}. In \bibinfo{booktitle}{\emph{3rd International Conference on Learning Representations, {ICLR} 2015, San Diego, CA, USA, May 7-9, 2015, Conference Track Proceedings}}, \bibfield{editor}{\bibinfo{person}{Yoshua Bengio} {and} \bibinfo{person}{Yann LeCun}} (Eds.).
\newblock


\bibitem[Kolkin et~al\mbox{.}(2019)]%
        {kolkin2019style}
\bibfield{author}{\bibinfo{person}{Nicholas Kolkin}, \bibinfo{person}{Jason Salavon}, {and} \bibinfo{person}{Gregory Shakhnarovich}.} \bibinfo{year}{2019}\natexlab{}.
\newblock \showarticletitle{Style transfer by relaxed optimal transport and self-similarity}. In \bibinfo{booktitle}{\emph{Proceedings of the IEEE/CVF Conference on Computer Vision and Pattern Recognition}}.
\newblock


\bibitem[Li and Wand(2016a)]%
        {LiW16}
\bibfield{author}{\bibinfo{person}{Chuan Li} {and} \bibinfo{person}{Michael Wand}.} \bibinfo{year}{2016}\natexlab{a}.
\newblock \showarticletitle{Combining Markov Random Fields and Convolutional Neural Networks for Image Synthesis}.
\newblock


\bibitem[Li and Wand(2016b)]%
        {li2016precomputed}
\bibfield{author}{\bibinfo{person}{Chuan Li} {and} \bibinfo{person}{Michael Wand}.} \bibinfo{year}{2016}\natexlab{b}.
\newblock \showarticletitle{Precomputed real-time texture synthesis with markovian generative adversarial networks}. In \bibinfo{booktitle}{\emph{European conference on computer vision}}. Springer, \bibinfo{pages}{702--716}.
\newblock


\bibitem[Li et~al\mbox{.}(2017)]%
        {li2017diversified}
\bibfield{author}{\bibinfo{person}{Yijun Li}, \bibinfo{person}{Chen Fang}, \bibinfo{person}{Jimei Yang}, \bibinfo{person}{Zhaowen Wang}, \bibinfo{person}{Xin Lu}, {and} \bibinfo{person}{Ming-Hsuan Yang}.} \bibinfo{year}{2017}\natexlab{}.
\newblock \showarticletitle{Diversified texture synthesis with feed-forward networks}. In \bibinfo{booktitle}{\emph{Proceedings of the IEEE Conference on Computer Vision and Pattern Recognition}}. \bibinfo{pages}{3920--3928}.
\newblock


\bibitem[Liao et~al\mbox{.}(2017)]%
        {Liao2017}
\bibfield{author}{\bibinfo{person}{Jing Liao}, \bibinfo{person}{Yuan Yao}, \bibinfo{person}{Lu Yuan}, \bibinfo{person}{Gang Hua}, {and} \bibinfo{person}{Sing~Bing Kang}.} \bibinfo{year}{2017}\natexlab{}.
\newblock \showarticletitle{Visual Attribute Transfer Through Deep Image Analogy}.
\newblock \bibinfo{journal}{\emph{ACM Trans. Graph.}} \bibinfo{volume}{36}, \bibinfo{number}{4}, Article \bibinfo{articleno}{120} (\bibinfo{date}{July} \bibinfo{year}{2017}), \bibinfo{numpages}{15}~pages.
\newblock
\showISSN{0730-0301}
\urldef\tempurl%
\url{https://doi.org/10.1145/3072959.3073683}
\showDOI{\tempurl}


\bibitem[Lin et~al\mbox{.}(2020)]%
        {lin2020tuigan}
\bibfield{author}{\bibinfo{person}{Jianxin Lin}, \bibinfo{person}{Yingxue Pang}, \bibinfo{person}{Yingce Xia}, \bibinfo{person}{Zhibo Chen}, {and} \bibinfo{person}{Jiebo Luo}.} \bibinfo{year}{2020}\natexlab{}.
\newblock \showarticletitle{Tuigan: Learning versatile image-to-image translation with two unpaired images}. In \bibinfo{booktitle}{\emph{European Conference on Computer Vision}}. Springer.
\newblock


\bibitem[Liu et~al\mbox{.}(2017)]%
        {10.5555/3294771.3294838}
\bibfield{author}{\bibinfo{person}{Ming-Yu Liu}, \bibinfo{person}{Thomas Breuel}, {and} \bibinfo{person}{Jan Kautz}.} \bibinfo{year}{2017}\natexlab{}.
\newblock \showarticletitle{Unsupervised Image-to-Image Translation Networks}. In \bibinfo{booktitle}{\emph{Proceedings of the 31st International Conference on Neural Information Processing Systems}} \emph{(\bibinfo{series}{NIPS'17})}. \bibinfo{publisher}{Curran Associates Inc.}
\newblock
\showISBNx{9781510860964}


\bibitem[Mahendran and Vedaldi(2014)]%
        {mahendran2014understanding}
\bibfield{author}{\bibinfo{person}{Aravindh Mahendran} {and} \bibinfo{person}{Andrea Vedaldi}.} \bibinfo{year}{2014}\natexlab{}.
\newblock \showarticletitle{Understanding deep image representations by inverting them}.
\newblock \bibinfo{journal}{\emph{2015 IEEE Conference on Computer Vision and Pattern Recognition (CVPR)}} (\bibinfo{year}{2014}), \bibinfo{pages}{5188--5196}.
\newblock
\urldef\tempurl%
\url{https://api.semanticscholar.org/CorpusID:206593185}
\showURL{%
\tempurl}


\bibitem[Mechrez et~al\mbox{.}(2018)]%
        {MechrezTZ18}
\bibfield{author}{\bibinfo{person}{Roey Mechrez}, \bibinfo{person}{Itamar Talmi}, {and} \bibinfo{person}{Lihi Zelnik{-}Manor}.} \bibinfo{year}{2018}\natexlab{}.
\newblock \showarticletitle{The Contextual Loss for Image Transformation with Non-aligned Data}.
\newblock


\bibitem[Melas-Kyriazi et~al\mbox{.}(2022)]%
        {melaskyriazi2022deep}
\bibfield{author}{\bibinfo{person}{Luke Melas-Kyriazi}, \bibinfo{person}{Christian Rupprecht}, \bibinfo{person}{Iro Laina}, {and} \bibinfo{person}{Andrea Vedaldi}.} \bibinfo{year}{2022}\natexlab{}.
\newblock \showarticletitle{Deep Spectral Methods: A Surprisingly Strong Baseline for Unsupervised Semantic Segmentation and Localization}. In \bibinfo{booktitle}{\emph{CVPR}}.
\newblock


\bibitem[Mokady et~al\mbox{.}(2022)]%
        {mokady2022selfdistilled}
\bibfield{author}{\bibinfo{person}{Ron Mokady}, \bibinfo{person}{Omer Tov}, \bibinfo{person}{Michal Yarom}, \bibinfo{person}{Oran Lang}, \bibinfo{person}{Inbar Mosseri}, \bibinfo{person}{Tali Dekel}, \bibinfo{person}{Daniel Cohen-Or}, {and} \bibinfo{person}{Michal Irani}.} \bibinfo{year}{2022}\natexlab{}.
\newblock \showarticletitle{Self-Distilled StyleGAN: Towards Generation from Internet Photos}. In \bibinfo{booktitle}{\emph{ACM SIGGRAPH 2022 Conference Proceedings}} \emph{(\bibinfo{series}{SIGGRAPH '22})}. \bibinfo{publisher}{Association for Computing Machinery}, Article \bibinfo{articleno}{50}, \bibinfo{numpages}{9}~pages.
\newblock
\showISBNx{9781450393379}
\urldef\tempurl%
\url{https://doi.org/10.1145/3528233.3530708}
\showDOI{\tempurl}


\bibitem[Naseer et~al\mbox{.}(2021)]%
        {naseer2021intriguing}
\bibfield{author}{\bibinfo{person}{Muzammal Naseer}, \bibinfo{person}{Kanchana Ranasinghe}, \bibinfo{person}{Salman Khan}, \bibinfo{person}{Munawar Hayat}, \bibinfo{person}{Fahad Khan}, {and} \bibinfo{person}{Ming-Hsuan Yang}.} \bibinfo{year}{2021}\natexlab{}.
\newblock \showarticletitle{Intriguing Properties of Vision Transformers}. In \bibinfo{booktitle}{\emph{Advances in Neural Information Processing Systems}}, \bibfield{editor}{\bibinfo{person}{A.~Beygelzimer}, \bibinfo{person}{Y.~Dauphin}, \bibinfo{person}{P.~Liang}, {and} \bibinfo{person}{J.~Wortman Vaughan}} (Eds.).
\newblock
\urldef\tempurl%
\url{https://openreview.net/forum?id=o2mbl-Hmfgd}
\showURL{%
\tempurl}


\bibitem[Nilsback and Zisserman(2008)]%
        {oxford102}
\bibfield{author}{\bibinfo{person}{Maria-Elena Nilsback} {and} \bibinfo{person}{Andrew Zisserman}.} \bibinfo{year}{2008}\natexlab{}.
\newblock \showarticletitle{Automated Flower Classification over a Large Number of Classes}. In \bibinfo{booktitle}{\emph{2008 Sixth Indian Conference on Computer Vision, Graphics Image Processing}}.
\newblock


\bibitem[Olah et~al\mbox{.}(2017)]%
        {olah2017feature}
\bibfield{author}{\bibinfo{person}{Chris Olah}, \bibinfo{person}{Alexander Mordvintsev}, {and} \bibinfo{person}{Ludwig Schubert}.} \bibinfo{year}{2017}\natexlab{}.
\newblock \showarticletitle{Feature Visualization}.
\newblock \bibinfo{journal}{\emph{Distill}} (\bibinfo{year}{2017}).
\newblock


\bibitem[Park et~al\mbox{.}(2020a)]%
        {park2020contrastive}
\bibfield{author}{\bibinfo{person}{Taesung Park}, \bibinfo{person}{Alexei~A Efros}, \bibinfo{person}{Richard Zhang}, {and} \bibinfo{person}{Jun-Yan Zhu}.} \bibinfo{year}{2020}\natexlab{a}.
\newblock \showarticletitle{Contrastive learning for unpaired image-to-image translation}. In \bibinfo{booktitle}{\emph{European Conference on Computer Vision}}. Springer.
\newblock


\bibitem[Park et~al\mbox{.}(2020b)]%
        {park2020swapping}
\bibfield{author}{\bibinfo{person}{Taesung Park}, \bibinfo{person}{Jun-Yan Zhu}, \bibinfo{person}{Oliver Wang}, \bibinfo{person}{Jingwan Lu}, \bibinfo{person}{Eli Shechtman}, \bibinfo{person}{Alexei~A. Efros}, {and} \bibinfo{person}{Richard Zhang}.} \bibinfo{year}{2020}\natexlab{b}.
\newblock \showarticletitle{Swapping Autoencoder for Deep Image Manipulation}. In \bibinfo{booktitle}{\emph{Advances in Neural Information Processing Systems}}.
\newblock


\bibitem[Paszke et~al\mbox{.}(2019)]%
        {NEURIPS2019PyTorch}
\bibfield{author}{\bibinfo{person}{Adam Paszke}, \bibinfo{person}{Sam Gross}, \bibinfo{person}{Francisco Massa}, \bibinfo{person}{Adam Lerer}, \bibinfo{person}{James Bradbury}, \bibinfo{person}{Gregory Chanan}, \bibinfo{person}{Trevor Killeen}, \bibinfo{person}{Zeming Lin}, \bibinfo{person}{Natalia Gimelshein}, \bibinfo{person}{Luca Antiga}, \bibinfo{person}{Alban Desmaison}, \bibinfo{person}{Andreas Kopf}, \bibinfo{person}{Edward Yang}, \bibinfo{person}{Zachary DeVito}, \bibinfo{person}{Martin Raison}, \bibinfo{person}{Alykhan Tejani}, \bibinfo{person}{Sasank Chilamkurthy}, \bibinfo{person}{Benoit Steiner}, \bibinfo{person}{Lu Fang}, \bibinfo{person}{Junjie Bai}, {and} \bibinfo{person}{Soumith Chintala}.} \bibinfo{year}{2019}\natexlab{}.
\newblock \showarticletitle{PyTorch: An Imperative Style, High-Performance Deep Learning Library}.
\newblock In \bibinfo{booktitle}{\emph{Advances in Neural Information Processing Systems 32}}, \bibfield{editor}{\bibinfo{person}{H.~Wallach}, \bibinfo{person}{H.~Larochelle}, \bibinfo{person}{A.~Beygelzimer}, \bibinfo{person}{F.~d\textquotesingle Alch\'{e}-Buc}, \bibinfo{person}{E.~Fox}, {and} \bibinfo{person}{R.~Garnett}} (Eds.). \bibinfo{publisher}{Curran Associates, Inc.}, \bibinfo{pages}{8024--8035}.
\newblock
\urldef\tempurl%
\url{http://papers.neurips.cc/paper/9015-pytorch-an-imperative-style-high-performance-deep-learning-library.pdf}
\showURL{%
\tempurl}


\bibitem[Ronneberger et~al\mbox{.}(2015)]%
        {ronneberger2015u}
\bibfield{author}{\bibinfo{person}{Olaf Ronneberger}, \bibinfo{person}{Philipp Fischer}, {and} \bibinfo{person}{Thomas Brox}.} \bibinfo{year}{2015}\natexlab{}.
\newblock \showarticletitle{U-net: Convolutional networks for biomedical image segmentation}. In \bibinfo{booktitle}{\emph{International Conference on Medical image computing and computer-assisted intervention}}. Springer.
\newblock


\bibitem[Shechtman and Irani(2007)]%
        {shechtman2007localselfsim}
\bibfield{author}{\bibinfo{person}{Eli Shechtman} {and} \bibinfo{person}{Michal Irani}.} \bibinfo{year}{2007}\natexlab{}.
\newblock \showarticletitle{Matching local self-similarities across images and videos}. In \bibinfo{booktitle}{\emph{CVPR}}.
\newblock


\bibitem[Shih et~al\mbox{.}(2014)]%
        {BarnesF14a}
\bibfield{author}{\bibinfo{person}{Yi{-}Chang Shih}, \bibinfo{person}{Sylvain Paris}, \bibinfo{person}{Connelly Barnes}, \bibinfo{person}{William~T. Freeman}, {and} \bibinfo{person}{Fr{\'{e}}do Durand}.} \bibinfo{year}{2014}\natexlab{}.
\newblock \showarticletitle{Style transfer for headshot portraits}.
\newblock \bibinfo{journal}{\emph{{ACM} Trans. Graph.}} (\bibinfo{year}{2014}).
\newblock


\bibitem[Shih et~al\mbox{.}(2013)]%
        {ShihPDF13}
\bibfield{author}{\bibinfo{person}{Yi{-}Chang Shih}, \bibinfo{person}{Sylvain Paris}, \bibinfo{person}{Fr{\'{e}}do Durand}, {and} \bibinfo{person}{William~T. Freeman}.} \bibinfo{year}{2013}\natexlab{}.
\newblock \showarticletitle{Data-driven hallucination of different times of day from a single outdoor photo}.
\newblock \bibinfo{journal}{\emph{{ACM} Trans. Graph.}} (\bibinfo{year}{2013}).
\newblock


\bibitem[Sim\'eoni et~al\mbox{.}(2021)]%
        {LOST}
\bibfield{author}{\bibinfo{person}{Oriane Sim\'eoni}, \bibinfo{person}{Gilles Puy}, \bibinfo{person}{Huy~V. Vo}, \bibinfo{person}{Simon Roburin}, \bibinfo{person}{Spyros Gidaris}, \bibinfo{person}{Andrei Bursuc}, \bibinfo{person}{Patrick P\'erez}, \bibinfo{person}{Renaud Marlet}, {and} \bibinfo{person}{Jean Ponce}.} \bibinfo{year}{2021}\natexlab{}.
\newblock \showarticletitle{Localizing Objects with Self-Supervised Transformers and no Labels}.
\newblock \bibinfo{journal}{\emph{Proceedings of the British Machine Vision Conference (BMVC)}}.
\newblock


\bibitem[Simonyan et~al\mbox{.}(2014)]%
        {simonyan2014deep}
\bibfield{author}{\bibinfo{person}{Karen Simonyan}, \bibinfo{person}{Andrea Vedaldi}, {and} \bibinfo{person}{Andrew Zisserman}.} \bibinfo{year}{2014}\natexlab{}.
\newblock \showarticletitle{Deep inside convolutional networks: Visualising image classification models and saliency maps}. In \bibinfo{booktitle}{\emph{In Workshop at International Conference on Learning Representations}}.
\newblock


\bibitem[Taigman et~al\mbox{.}(2017)]%
        {DBLP:conf/iclr/TaigmanPW17}
\bibfield{author}{\bibinfo{person}{Yaniv Taigman}, \bibinfo{person}{Adam Polyak}, {and} \bibinfo{person}{Lior Wolf}.} \bibinfo{year}{2017}\natexlab{}.
\newblock \showarticletitle{Unsupervised Cross-Domain Image Generation}. In \bibinfo{booktitle}{\emph{5th International Conference on Learning Representations, {ICLR} 2017, Toulon, France, April 24-26, 2017, Conference Track Proceedings}}. \bibinfo{publisher}{OpenReview.net}.
\newblock
\urldef\tempurl%
\url{https://openreview.net/forum?id=Sk2Im59ex}
\showURL{%
\tempurl}


\bibitem[Ulyanov et~al\mbox{.}(2016)]%
        {ulyanov2016texture}
\bibfield{author}{\bibinfo{person}{Dmitry Ulyanov}, \bibinfo{person}{Vadim Lebedev}, \bibinfo{person}{Andrea Vedaldi}, {and} \bibinfo{person}{Victor~S Lempitsky}.} \bibinfo{year}{2016}\natexlab{}.
\newblock \showarticletitle{Texture networks: Feed-forward synthesis of textures and stylized images.}. In \bibinfo{booktitle}{\emph{ICML}}, Vol.~\bibinfo{volume}{1}. \bibinfo{pages}{4}.
\newblock


\bibitem[Ulyanov et~al\mbox{.}(2018)]%
        {UlyanovVL17}
\bibfield{author}{\bibinfo{person}{Dmitry Ulyanov}, \bibinfo{person}{Andrea Vedaldi}, {and} \bibinfo{person}{Victor Lempitsky}.} \bibinfo{year}{2018}\natexlab{}.
\newblock \showarticletitle{Deep Image Prior}. In \bibinfo{booktitle}{\emph{Proceedings of the IEEE Conference on Computer Vision and Pattern Recognition (CVPR)}}.
\newblock


\bibitem[Wang et~al\mbox{.}(2018)]%
        {Fast_Photographic}
\bibfield{author}{\bibinfo{person}{Li Wang}, \bibinfo{person}{Nan Xiang}, \bibinfo{person}{Xiaosong Yang}, {and} \bibinfo{person}{Jianjun Zhang}.} \bibinfo{year}{2018}\natexlab{}.
\newblock \showarticletitle{Fast Photographic Style Transfer Based on Convolutional Neural Networks}. In \bibinfo{booktitle}{\emph{Proceedings of Computer Graphics International 2018}} \emph{(\bibinfo{series}{CGI 2018})}. \bibinfo{publisher}{Association for Computing Machinery}, \bibinfo{address}{New York, NY, USA}.
\newblock


\bibitem[Wang et~al\mbox{.}(2022)]%
        {wang2022tokencut}
\bibfield{author}{\bibinfo{person}{Yangtao Wang}, \bibinfo{person}{Xi Shen}, \bibinfo{person}{Shell~Xu Hu}, \bibinfo{person}{Yuan Yuan}, \bibinfo{person}{James~L. Crowley}, {and} \bibinfo{person}{Dominique Vaufreydaz}.} \bibinfo{year}{2022}\natexlab{}.
\newblock \showarticletitle{Self-supervised Transformers for Unsupervised Object Discovery using Normalized Cut}. In \bibinfo{booktitle}{\emph{Conference on Computer Vision and Pattern Recognition}}. \bibinfo{address}{New Orleans, LA, USA}.
\newblock


\bibitem[Wilmot et~al\mbox{.}(2017)]%
        {Wilmot2017StableAC}
\bibfield{author}{\bibinfo{person}{Pierre Wilmot}, \bibinfo{person}{Eric Risser}, {and} \bibinfo{person}{Connelly Barnes}.} \bibinfo{year}{2017}\natexlab{}.
\newblock \showarticletitle{Stable and Controllable Neural Texture Synthesis and Style Transfer Using Histogram Losses}.
\newblock \bibinfo{journal}{\emph{ArXiv}} (\bibinfo{year}{2017}).
\newblock


\bibitem[Xu et~al\mbox{.}(2020)]%
        {XuWFSZ20}
\bibfield{author}{\bibinfo{person}{Zhongyou Xu}, \bibinfo{person}{Tingting Wang}, \bibinfo{person}{Faming Fang}, \bibinfo{person}{Yun Sheng}, {and} \bibinfo{person}{Guixu Zhang}.} \bibinfo{year}{2020}\natexlab{}.
\newblock \showarticletitle{Stylization-Based Architecture for Fast Deep Exemplar Colorization}.
\newblock


\bibitem[Yi et~al\mbox{.}(2017)]%
        {8237572}
\bibfield{author}{\bibinfo{person}{Zili Yi}, \bibinfo{person}{Hao Zhang}, \bibinfo{person}{Ping Tan}, {and} \bibinfo{person}{Minglun Gong}.} \bibinfo{year}{2017}\natexlab{}.
\newblock \showarticletitle{DualGAN: Unsupervised Dual Learning for Image-to-Image Translation}. In \bibinfo{booktitle}{\emph{2017 IEEE International Conference on Computer Vision (ICCV)}}.
\newblock


\bibitem[Yoo et~al\mbox{.}(2019)]%
        {Yoo_2019_ICCV}
\bibfield{author}{\bibinfo{person}{Jaejun Yoo}, \bibinfo{person}{Youngjung Uh}, \bibinfo{person}{Sanghyuk Chun}, \bibinfo{person}{Byeongkyu Kang}, {and} \bibinfo{person}{Jung-Woo Ha}.} \bibinfo{year}{2019}\natexlab{}.
\newblock \showarticletitle{Photorealistic Style Transfer via Wavelet Transforms}. In \bibinfo{booktitle}{\emph{Proceedings of the IEEE/CVF International Conference on Computer Vision (ICCV)}}.
\newblock


\bibitem[Zhang et~al\mbox{.}(2018)]%
        {lpips}
\bibfield{author}{\bibinfo{person}{Richard Zhang}, \bibinfo{person}{Phillip Isola}, \bibinfo{person}{Alexei~A Efros}, \bibinfo{person}{Eli Shechtman}, {and} \bibinfo{person}{Oliver Wang}.} \bibinfo{year}{2018}\natexlab{}.
\newblock \showarticletitle{The Unreasonable Effectiveness of Deep Features as a Perceptual Metric}. In \bibinfo{booktitle}{\emph{CVPR}}.
\newblock


\bibitem[Zhou et~al\mbox{.}(2018)]%
        {zhou2018semantic_seg}
\bibfield{author}{\bibinfo{person}{Bolei Zhou}, \bibinfo{person}{Hang Zhao}, \bibinfo{person}{Xavier Puig}, \bibinfo{person}{Tete Xiao}, \bibinfo{person}{Sanja Fidler}, \bibinfo{person}{Adela Barriuso}, {and} \bibinfo{person}{Antonio Torralba}.} \bibinfo{year}{2018}\natexlab{}.
\newblock \showarticletitle{Semantic understanding of scenes through the ade20k dataset}.
\newblock \bibinfo{journal}{\emph{International Journal on Computer Vision}} (\bibinfo{year}{2018}).
\newblock


\bibitem[Zhu et~al\mbox{.}(2017)]%
        {8237506}
\bibfield{author}{\bibinfo{person}{Jun-Yan Zhu}, \bibinfo{person}{Taesung Park}, \bibinfo{person}{Phillip Isola}, {and} \bibinfo{person}{Alexei~A. Efros}.} \bibinfo{year}{2017}\natexlab{}.
\newblock \showarticletitle{Unpaired Image-to-Image Translation Using Cycle-Consistent Adversarial Networks}. In \bibinfo{booktitle}{\emph{2017 IEEE International Conference on Computer Vision (ICCV)}}.
\newblock


\end{thebibliography}

\newpage

\appendix

\section{Architecture}
\label{sec:appendix-architecture}

\subsection{Splice Generator Architecture}
\label{sec:appendix-splice-architecture}

We base our generator $G_\theta$ network on a \texttt{U-Net} architecture~\cite{ronneberger2015u}, with a 5-layer encoder and a symmetrical decoder. All layers comprise $3\!\times\!3$ Convolutions, followed by \texttt{BatchNorm}, and \texttt{LeakyReLU} activation. The encoder's channels dimensions are $[3\rightarrow16\rightarrow32\rightarrow64\rightarrow128\rightarrow128]$ (the decoder follows a reversed order).
In each level of the encoder, we add an additional $1\!\times\!1$ Convolution layer and concatenate the output features to the corresponding level of the decoder. Lastly, we add a $1\!\times\!1$ Convolution layer followed by \texttt{Sigmoid} activation to get the final RGB output. 

\vspace{-3.4mm}
\subsection{SpliceNet Generator Architecture}
\label{sec:appendix-splicenet-architecture}

We design our feed-forward model $F_\theta$ based on a \texttt{U-Net} architecture \cite{ronneberger2015u}. The input image is first passed through a $1\!\times\!1$ convolutional layer with $32$ output channels. The output is then passed through a 5-layer encoder with channel dimensions of $[64 \rightarrow 128 \rightarrow 256 \rightarrow 512 \rightarrow 1024]$, followed by a symmetrical decoder. Each layer of the encoder is a downsampling residual block that is comprised of two consecutive $3\!\times\!3$ convolutions and a $1\!\times\!1$ convolution for establishing the residual connection. The decoder consists of upsampling residual blocks with a similar composition of convolutions and residual connection as in the encoder. In the decoder, the weights of the $3\!\times\!3$ convolutions are modulated with the input [CLS] token. In each layer of the encoder, in order to establish the skip connections to the decoder, the output features are passed through a resolution-preserving residual block, which is concatenated to the input of the decoder layer. The residual blocks in the skip connections have a similar composition of convolutions and modulations as the decoder residual blocks. Finally, the output of the last decoder layer is passed through a modulated $1\!\times\!1$ convolutional layer followed by a \texttt{Sigmoid} activation that produces the final RGB output.  \texttt{LeakyReLU} is used as an activation function in all the convolutional layers of the model.

Our mapping network $M$ is a 2-layer MLP that takes as input the [CLS] token $t_{\mathcls} \in\mathbb{R}^{768}$ extracted from DINO-ViT, and passes it through one hidden layer and an output layer, both with output dimensions of $768$ and with \texttt{GELU} activations. Following \cite{stylegan2}, for each modulated convolution in the feed-forward model, an affine transformation is learned that maps the output of the mapping network $M$ to a vector used for modulating the weights.

\vspace{-2mm}
\section{ViT Feature Extractor Architecture}
\label{sec:appendix-extractor-architecture}

As described in Sec. 3, we leverage a pre-trained ViT model (\dinovit~\cite{dino}) trained in a self-supervised manner as a feature extractor. We use the 12 layer pretrained model in the $8\!\times\!8$ patches configuration (\texttt{ViT-B/8}), downloaded from the  \href{https://github.com/facebookresearch/dino}{official implementation at GitHub}.

\section{Training Details}
\label{sec:appendix-training}

We implement our framework in PyTorch \cite{NEURIPS2019PyTorch}. We optimize our full objective (Eq.~4, Sec.~3.3), with relative weights: $\alpha=0.1$, $\beta=0.1$ for Splice, and $\alpha=2$, $\beta=0.1$ for SpliceNet. We use the Adam optimizer~\cite{DBLP:journals/corr/KingmaB14} with a constant learning rate of $\lambda=2\cdot 10^{-3}$ and with hyper-parameters $\beta_1 = 0$, $\beta_2 = 0.99$. Each batch contains {$\{\tilde{I_s}, \tilde{I_t}\}$}, the augmented views of the source structure image and the target appearance image respectively. For Splice, every 75 iterations, we add {$\{I_s$, $I_t\}$} to the batch (i.e., do not apply augmentations). All the images (both input and generated) are resized down to $224$[pix] (maintaining aspect ratio) using bicubic interpolation, before extracting \dinovit features for estimating the losses. The test-time training of Splice on an input image pair of size $512\!\times\!512$ takes $\sim\!20$ minutes to train on a single GPU (Nvidia RTX 6000) for a total of 2000 iterations.

\section{Data Augmentations}
\label{sec:appendix-data-aug}

At each training step, given an input pair {$\{I_s$, $I_t\}$}, we apply on them the following random augmentations:
Augmentations to the source structure image $I_s$:
\begin{itemize}
  \item cropping: we uniformly sample a {NxN} crop; N is between 95\% - 100\% of the height of $I_s$ (for SpliceNet, we fix N=95\%)
  \item horizontal-flipping, applied in probability {p=0.5}.
  \item color jittering: we jitter the brightness, contrast, saturation and hue of the image in probability p, where {p=0.5} for Splice and {p=0.2} for SpliceNet, 
  \item Gaussian blurring: we apply a Gaussian blurring {3x3} filter ($\sigma$ is uniformly sampled between 0.1-2.0) in probability p, where {p=0.5} for Splice and {p=0.1} for SpliceNet, 
\end{itemize}
Augmentations to the target appearance image $I_t$:
\begin{itemize}
  \item cropping: we uniformly sample a {NxN}; N is between 95\% - 100\% of the height of $I_t$ (for SpliceNet, we fix N=95\%).
  \item horizontal-flipping, applied in probability {p=0.5}.

\end{itemize}

\end{document}